\documentclass[USenglish,onecolumn]{article}
\usepackage{textcomp}

\usepackage{fix-cm}
\usepackage[utf8]{inputenc}
\usepackage[big]{dgruyter}

\makeatletter
\providecommand*{\toclevel@schapter}{0}
\makeatother
\usepackage{amsmath,amssymb,amsfonts}
\allowdisplaybreaks
\usepackage{booktabs}
\usepackage{multirow}
\usepackage{xcolor}
\usepackage{graphicx}
\graphicspath{{figures/}{logos/}}
\usepackage{float}
\usepackage{array}
\usepackage{subcaption}
\DeclareMathOperator*{\argmax}{argmax}
\DeclareMathOperator*{\argmin}{argmin}

\begin{document}

\articletype{Research Article{\hfill}Open Access}

  \author*[1]{Marco Ruiz}

  \author[2]{Miguel Arana-Catania}

  \author[3]{David R. Ardila}

  \author[4]{Rodrigo Ventura}

  \affil[1]{Institute for Systems and Robotics, Instituto Superior T\'{e}cnico, 1049-001 Lisbon, Portugal; E-mail: marco.rueda@tecnico.ulisboa.pt}

  \affil[2]{Digital Scholarship at Oxford, University of Oxford, Oxford, OX1 3BG, United Kingdom}

  \affil[3]{Jet Propulsion Laboratory, California Institute of Technology, Pasadena, CA 91109, USA}

  \affil[4]{Institute for Systems and Robotics, Instituto Superior T\'{e}cnico, 1049-001 Lisbon, Portugal}

  \title{\huge Causal-Audit: A Framework for Risk Assessment of Assumption Violations in Time-Series Causal Discovery}

  \runningtitle{Causal-Audit}

\begin{abstract}
{Time-series causal discovery methods rely on assumptions such as stationarity, regular sampling, and bounded temporal dependence. When these assumptions are violated, structure learning can produce confident but misleading causal graphs without warning. We introduce Causal-Audit, a framework that formalizes assumption validation as calibrated risk assessment. The framework computes effect-size diagnostics across five assumption families (stationarity, irregularity, persistence, nonlinearity, and confounding proxies), aggregates them into four calibrated risk scores with uncertainty intervals, and applies an abstention-aware decision policy that recommends methods (e.g., PCMCI+, VAR-based Granger causality) only when evidence supports reliable inference. The semi-automatic diagnostic stage can also be used independently for structured assumption auditing in individual studies. Evaluation on a synthetic atlas of 500 data-generating processes (DGPs) spanning 10 violation families demonstrates well-calibrated risk scores (AUROC \(> 0.95\)), a 62\% false positive reduction among recommended datasets, and 78\% abstention on severe-violation cases. On 21 external evaluations from TimeGraph (18 categories) and CausalTime (3 domains), recommend-or-abstain decisions are consistent with benchmark specifications in all cases. An open-source implementation of our framework is available.}
\end{abstract}


  \journalname{Journal of Causal Inference}
%\DOI{DOI}
  \startpage{1}
%  \received{..}
%  \revised{..}
%  \accepted{..}

  \journalyear{2026}
  \journalvolume{1}
%  \journalissue{1}

%% ===================================================================
\maketitle

\section{Introduction}
\label{sec:introduction}
%% ===================================================================

\begin{figure}[ht!]
\centering
\includegraphics[width=0.99\textwidth]{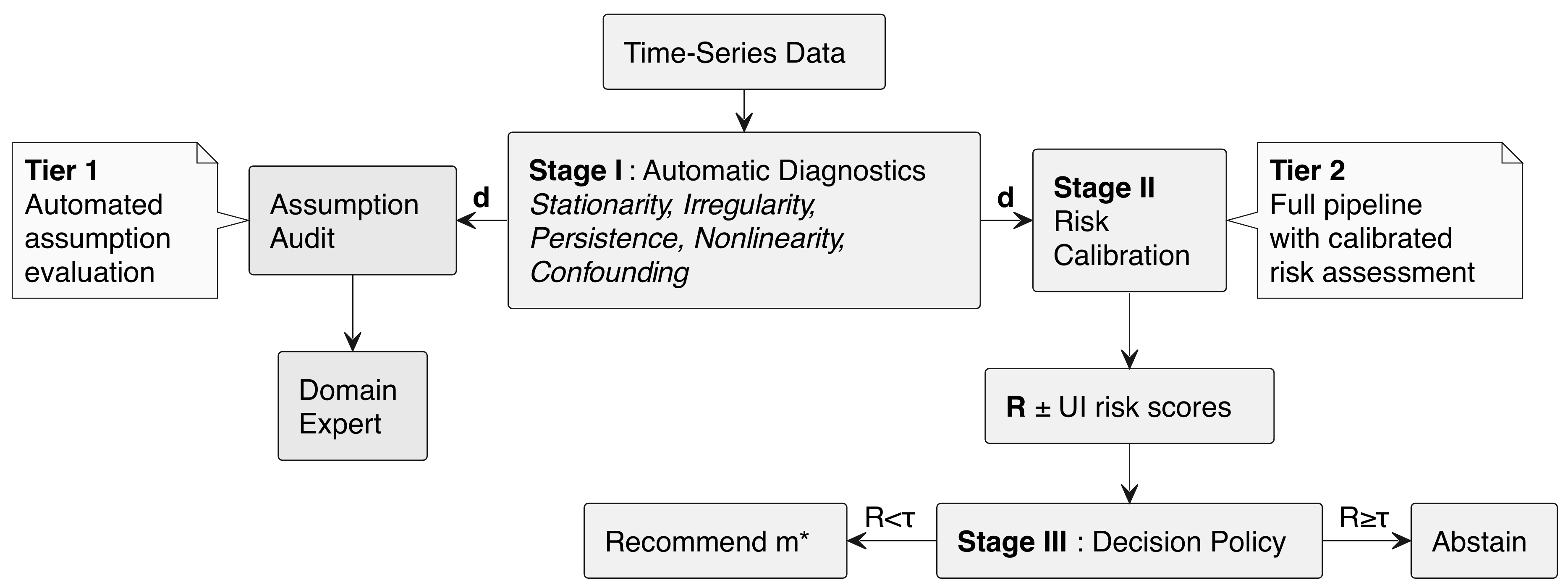}
\caption{Framework overview. Tier~1 (Stage~I alone) provides automatic diagnostics \textit{d} across five assumption families for expert-guided assumption auditing. Tier~2 (Stages~I--III) adds calibrated risk estimation with uncertainty intervals and an abstention-aware decision policy that recommends using or abstaining from a method~\(m^*\) according to its risk score \textit{R}.}
\label{fig:framework-overview}
\end{figure}

Causal discovery from observational time series aims to distinguish genuine causal relationships from spurious associations \cite{pearl2009,runge2023review}. The reliability of the discovered structures depends on whether the data satisfy the assumptions of each algorithm. Cliff et al.\ \cite{cliff2020} show that autocorrelation violations can inflate false positive rates to 100\%, and Granger and Newbold \cite{granger1974} established that non-stationary processes generate spurious associations. When violations go undetected, algorithms produce spurious causal graphs without warning \cite{peters2017}. We refer to this phenomenon as \emph{silent failure}: the method returns a result without any indication that the output may be unreliable.

These concerns are particularly relevant in climate science \cite{runge2019}, neuroscience \cite{friston2014}, epidemiology \cite{shojaie2022}, and economics \cite{stock2001}. For example, daily temperature readings are strongly autocorrelated, so a year of daily observations may contain only a few dozen effectively independent samples. A causal discovery algorithm that treats all 365 observations as independent will report links that are artifacts of the inflated sample size. Such failures are silent, and the algorithm does not provide warning that discovered edges may be spurious.

Despite the availability of causal discovery tools such as Tigramite \cite{runge2019} and causal-learn\cite{squires2023}, no systematic tools exist for assessing whether time-series data satisfy the required assumptions \emph{before} discovery is attempted. Existing diagnostic tools provide raw statistics without translating them into actionable risk estimates or aggregating evidence across diagnostic dimensions.

This gap motivates the central question: \emph{Given a time-series dataset and candidate causal discovery methods, how can we quantitatively assess the risk that assumption violations will lead to unreliable results, and how can this assessment inform principled decisions about method selection or abstention?} Causal-Audit\footnote{\url{https://github.com/marcoruizrueda/causal-audit}} addresses this question through a three-stage pipeline (Figure~\ref{fig:framework-overview}), where diagnostic auditing quantifies assumption violations, risk estimation transforms these quantifications into calibrated failure probabilities, and a decision policy determines whether the causal discovery methods can be applied, based on expected utility.

The decision-theoretic formulation builds on Berger \cite{berger1985}. A utility function over action-outcome pairs encodes asymmetric consequences: publishing spurious causal claims incurs a large penalty, whereas abstention incurs only a modest opportunity cost. The optimal action maximizes expected utility, recommending to proceed only when the estimated failure probability is sufficiently low (Section~\ref{sec:problem-formulation}).

Two usage tiers are supported. Stage~I alone provides structured assumption auditing for researchers who prefer direct diagnostic interpretation, documenting data limitations to support transparent reporting. The full pipeline (Stages~I through III) adds calibrated risk estimation and abstention-aware decision support for researchers processing large dataset collections or seeking automated quality control.

The current implementation calibrates risk scores and decision thresholds for two methods: PCMCI+ \cite{runge2020}, a constraint-based method that identifies lagged and contemporaneous causal links via momentary conditional independence, and VAR-based Granger causality \cite{granger1969,shojaie2022}, a regression-based method that tests whether past values of one variable improve prediction of another. These methods were selected because they represent the two dominant paradigms in time-series causal discovery (constraint-based and regression-based), have complementary assumption profiles, and are among the most frequently applied in practice. The diagnostic stage (Stage~I) is method-agnostic; the method-specific components are the risk calibration (Stage~II) and the decision thresholds (Stage~III), which can be extended to additional methods as calibration data become available.

An open-source implementation and the benchmark dataset are available at \url{https://github.com/marcoruizrueda/causal-audit}

\subsection{Contributions}

We introduce Causal-Audit, a framework that formalizes pre-discovery assumption validation as calibrated risk assessment. The contributions of our work are the following:

(1) we propose novel risk profiles \(\mathbf{R} \in [0,1]^4\) that quantify assumption-violation probabilities, with decision-theoretic abstention thresholds derived from explicit utility functions;

(2) we develop a three-stage pipeline that transforms effect-size diagnostics into calibrated risk scores with uncertainty quantification, and perform extensive analysis, achieving calibration slopes of 0.96 to 1.05, expected calibration error (ECE) below 0.05, and AUROC exceeding 0.95;

(3) we design and evaluate an abstention-aware decision policy that reduces false positive rates by 62\% among recommended datasets and abstains on 78\% of severe-violation cases;

(4) we perform validation on external benchmarks (TimeGraph, CausalTime) with decisions consistent with benchmark specifications on all 21 evaluations, demonstrating generalization beyond the training distribution; and

(5) we publish the Synthetic DGP Atlas, a novel benchmark of 500 datasets across 10 violation families, together with an open-source implementation of the full framework.

%% ===================================================================
\section{Related Work}
\label{sec:related-work}
%% ===================================================================

Causal-Audit draws on time-series causal discovery methods, empirical studies of assumption violations, calibration theory, and selective prediction. The gap it addresses arises because these lines of work remain disconnected: methods exist, violations are documented, and calibration theory is mature, yet no system integrates diagnostic evidence into pre-discovery risk assessment with formal abstention criteria.

\subsection{Causal Discovery Methods for Time Series}

Structure learning from observational data relies on constraint-based, score-based, or functional causal model paradigms \cite{glymour2019review,guo2020survey,gong2024temporal}, each requiring assumptions such as causal sufficiency, faithfulness, and the Markov condition \cite{spirtes2001,vowels2022dags}. The time-series setting introduces additional requirements, including temporal precedence and autocorrelation \cite{runge2023review}. PCMCI \cite{runge2019} tests conditional independences among lagged variables, while PCMCI+ \cite{runge2020} extends this to contemporaneous links via momentary conditional independence. LPCMCI \cite{gerhardus2020} relaxes causal sufficiency by outputting partial ancestral graphs with bidirected edges indicating latent confounders.

Score-based methods optimize penalized likelihood objectives. DYNOTEARS \cite{pamfil2020} extends NOTEARS \cite{zheng2018} to time series using differentiable acyclicity constraints, though such methods often impose distributional constraints that degrade performance when violated. Functional causal models exploit non-Gaussianity for identifiability: VAR-LiNGAM \cite{hyvarinen2008} applies independent component analysis to VAR residuals, with nonlinear extensions such as TCDF \cite{nauta2019}. Information-theoretic approaches, including transfer entropy \cite{schreiber2000} and its partial variant \cite{kugiumtzis2013}, avoid parametric assumptions but require large samples.

Recent methods target specific violations: CD-NOD \cite{gong2015} handles changing causal structures across environments, Huang et al.\ \cite{huang2020} address causal discovery from heterogeneous and nonstationary data by exploiting distribution shifts as a source of identifiability, and SDCI \cite{balsells2025} addresses regime changes in conditionally stationary series. Comparative evaluations reveal heterogeneous robustness profiles \cite{yi2025,montagna2023,montagna2023benchmark}, yet no method dominates universally, making method selection dependent on data characteristics.

Montagna et al.\ \cite{montagna2023benchmark} provide the first systematic robustness benchmark for iid causal discovery, evaluating eleven methods under six types of assumption violations (confounding, measurement error, unfaithfulness, autoregression, post-nonlinear models, and non-Gaussianity) across more than 60,000 synthetic datasets. Their analysis reveals that score-matching methods are surprisingly robust, but the benchmark is restricted to iid data and explicitly excludes time-series algorithms. TCD-Arena \cite{tcdarena2026} extends robustness benchmarking to the time-series setting, evaluating eight methods against 27 assumption violations over more than 50 million runs. Each method displays a distinct robustness profile: constraint-based approaches tolerate moderate non-stationarity yet deteriorate under latent confounding, whereas score-based methods show the opposite pattern. Performance degradation is not proportional to violation severity; instead, each method tolerates violations up to a characteristic threshold beyond which accuracy drops steeply. Ensemble strategies attenuate some weaknesses, but no fixed combination eliminates them. Both benchmarks, however, evaluate how well discovery methods recover graphs when assumptions are violated. They do not evaluate pre-discovery diagnostic tools that assess whether a dataset satisfies methodological assumptions before structure learning is attempted. Because both operate on synthetic data with known ground truth graphs, they can quantify method fragilities precisely, yet they offer no procedure for estimating the violation profile of a new, real-world dataset or for selecting a method on that basis. Causal-Audit fills this gap by inferring the violation profile from the data and mapping it to a calibrated method recommendation or an abstention decision.

\subsection{Assumption Violations and Pre-Discovery Assessment}

\begin{figure}[ht!]
  \centering
  \includegraphics[width=0.84\textwidth]{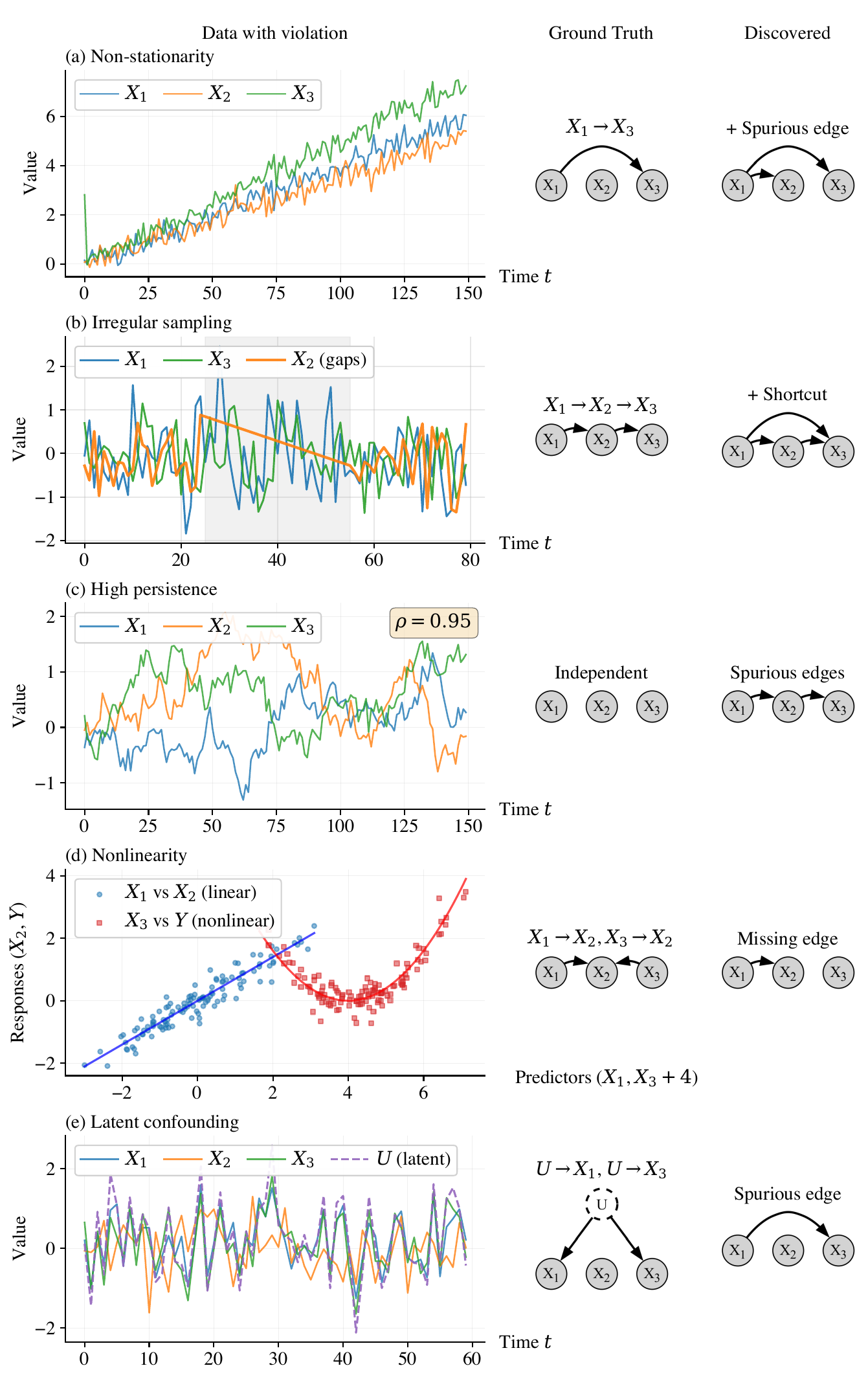}
  \caption{%
    Assumption violations in causal discovery.
    Each row shows data violating an assumption (left) and the resulting causal graph with true and erroneous edges (right; see legend).
  }
  \label{fig:assumptions}
\end{figure}

Stationarity requires that the joint distribution of the process remains constant over time. When trends, structural breaks, or regime changes violate this condition, regression-based and conditional independence methods detect shared non-stationary dynamics rather than genuine causal links, producing spurious associations \cite{granger1974,phillips1986,balsells2025}. Irregular sampling and missing data pose a different threat: temporal gaps distort the lag structure on which causal discovery relies, and non-random missingness introduces selection bias that cannot be corrected without additional assumptions \cite{rehfeld2011,little2019,ding2018}. Strong temporal persistence, where successive observations carry nearly the same information, reduces the effective number of independent data points, so that standard significance tests overstate the evidence for causal links \cite{bayley1946,thiebaux1984,cliff2020}. Nonlinear relationships create the opposite problem: linear conditional independence tests may fail to detect genuine dependencies, leading to missed causal edges \cite{peters2017,shah2020}. Finally, high multicollinearity and parameter instability destabilize coefficient estimates and signal model misspecification, compromising the reliability of any structure learned from the data \cite{hoerl1970,white1980}. Each assumption family produces distinct statistical artifacts when ignored (Table~\ref{tab:consequences}). Figure~\ref{fig:assumptions} illustrates these failure modes in synthetic multivariate time series.

\begin{table}[ht!]
\centering
\caption{Assumption families and consequences of violations. Causal-Audit computes diagnostics for all five families and aggregates four (all except nonlinearity) into calibrated risk scores.
\label{tab:consequences}}
\begin{tabular}{llll}
\toprule
\textbf{Family} & \textbf{Assumption} & \textbf{Consequence if Violated} & \textbf{Error} \\
\midrule
Stationarity & No trends or regime changes & Spurious links from trends & FP \\
Irregularity & No gaps or missing values & Selection bias, incorrect lags & Bias \\
Persistence & Low autocorrelation & Inflated significance & FP \\
Nonlinearity & Linear relationships & Missed nonlinear effects & FN \\
Confounding proxies & Stable model coefficients & Spurious links from instability & FP \\
\bottomrule
\end{tabular}
\end{table}

Despite the severity of these violations, systematic pre-discovery assessment remains rare. The violation dimensions in Table~\ref{tab:consequences} overlap substantially with the 27 violation types catalogued by Stein et al.\ \cite{tcdarena2026}, who parameterize each violation within a synthetic data-generating process and measure accuracy degradation as severity increases. Their approach requires ground truth graphs and controlled violation injection, which is feasible for benchmarking but not for real-world datasets where the violation profile must be inferred from the data alone. Our framework complements this by replacing controlled violation injection with statistical diagnostic tests (ADF, KPSS, Ljung-Box, Bai-Perron, Little's MCAR, among others) that estimate violation presence and severity directly from observed data, and by extending coverage to irregularity and missing-data violations outside TCD-Arena's VAR-based simulation scope.

Individual tests exist for stationarity (ADF \cite{dickey1979}, KPSS \cite{kwiatkowski1992}), structural breaks \cite{bai2003}, missingness patterns \cite{little1988}, and autocorrelation \cite{ljung1978}, but these diagnostics are typically applied in isolation, yielding binary decisions rather than composite risk assessments.

Existing uncertainty quantification for causal discovery is \emph{post hoc}: Bayesian approaches infer posterior distributions over graphs \cite{heckerman1995,friedman2003}, and frequentist alternatives assess edge stability via bootstrapping \cite{ramsey2018,debeire2024}. These methods assume a causal graph has already been estimated. Our work is \emph{pre hoc}, quantifying the risk that running a method will produce unreliable results \emph{before} discovery. We adopt isotonic regression \cite{zadrozny2002,barlow1972} for calibration because it makes no parametric assumptions about the score-to-probability mapping, accommodates arbitrary monotonic relationships, and preserves the ROC curve convex hull \cite{berta2024,dawid1982,degroot1983}. Selective prediction, which permits models to withhold predictions on uncertain inputs \cite{chow1970,elyaniv2010}, provides the decision-theoretic foundation for our abstention mechanism.

%% ===================================================================
\section{Problem Formulation}
\label{sec:problem-formulation}
%% ===================================================================

\subsection{Setting and Notation}
\label{sec:setting-notation}

\begin{figure}[ht!]
\centering
\includegraphics[width=0.99\textwidth]{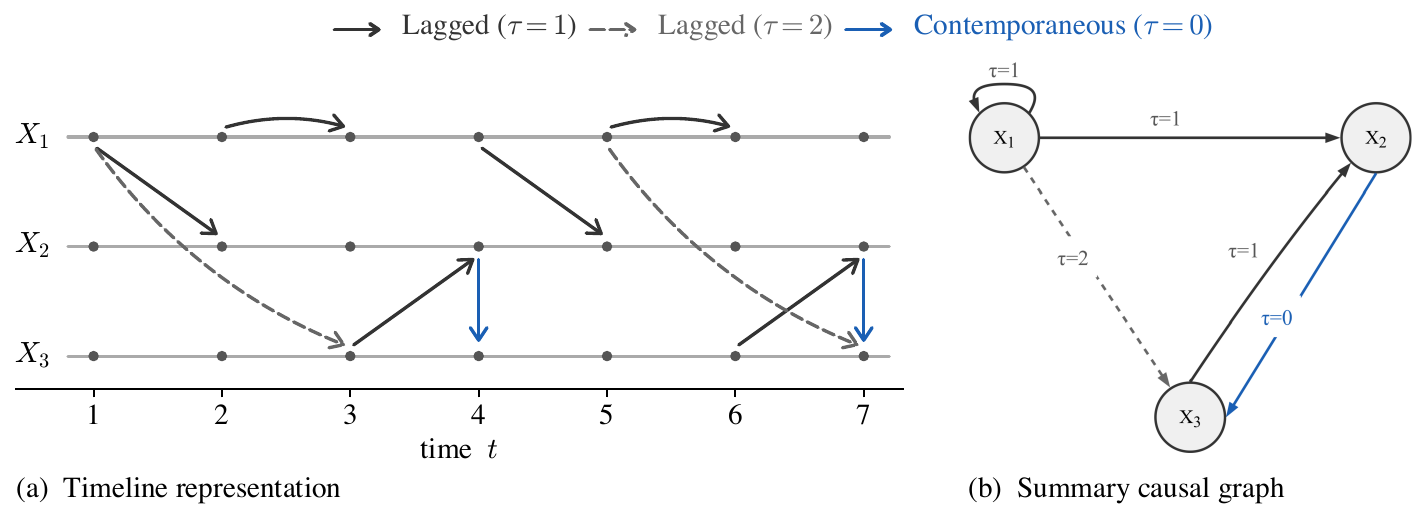}
\caption{Time series causal graph \(\mathcal{G} = (V, E)\) for \(N = 3\) variables with \(\tau_{\max} = 2\). (a)~Timeline representation: arrows between variable timelines encode causal effects; horizontal span equals the lag~\(\tau\). Each edge repeats at every time step (stationarity). (b)~Summary causal graph: each directed edge is annotated with its lag, corresponding to the triple \((i, j, \tau) \in E\).}
\label{fig:temporal-dag}
\end{figure}

Let \(\mathbf{X} = (\mathbf{x}_1, \ldots, \mathbf{x}_T)^\top \in \mathbb{R}^{T \times N}\) denote a multivariate time series with \(T\) observations of \(N\) variables. The underlying causal structure is encoded by a directed graph \(\mathcal{G} = (V, E)\) with edges \(E \subseteq V \times V \times \{0, \ldots, \tau_{\max}\}\) encoding lagged causal relationships. Figure~\ref{fig:temporal-dag} illustrates this representation for \(N = 3\) variables and \(\tau_{\max} = 2\). Panel~(a) unfolds the graph across time, where each horizontal line tracks one variable and arrows between time points encode causal effects with horizontal span proportional to the lag~\(\tau\). The same edges reappear at every time step, reflecting the stationarity assumption. Panel~(b) compresses this repeating structure into a summary graph in which each directed edge carries its lag annotation, so that the triple \((i, j, \tau) \in E\) can be read directly.

A catalog of causal discovery methods \(\mathcal{M} = \{m_1, \ldots, m_K\}\) is available, where each method \(m_k\) has documented assumptions \(\mathcal{A}_k\). The objective is to assess whether \(\mathbf{X}\) satisfies the prerequisites of each candidate method, producing: (i) a risk profile \(\mathbf{R} = (R_{\mathrm{nonstat}}, R_{\mathrm{irreg}}, R_{\mathrm{persist}}, R_{\mathrm{confound}}) \in [0,1]^4\); (ii) uncertainty intervals \([R_k^{-}, R_k^{+}]\); and (iii) a decision \(d \in \mathcal{M} \cup \{Abstain\}\).

The framework organizes diagnostics into four risk dimensions, each corresponding to assumptions that can cause method failure through distinct mechanisms. Nonlinearity, while important, is treated as a method-specific assumption and computed as a separate diagnostic rather than a standalone risk dimension, since its consequences depend strongly on the functional form assumed by each method.

\emph{Nonstationarity risk} \(R_{\mathrm{nonstat}}\) quantifies threats from non-stationary dynamics. Stationarity requires the joint distribution of the process to remain constant over time. Unit roots introduce stochastic trends that generate spurious correlations between otherwise independent series \cite{granger1974,phillips1986}. Structural breaks shift the data-generating process at discrete points, so that conditional independence relations estimated over the full sample conflate distinct regimes. Time-varying parameters produce a similar effect gradually, as the coefficients governing causal relationships drift across the observation window. In all three cases, constraint-based methods evaluate a single set of dependence relations that does not hold uniformly over the sample, yielding both false positive and false negative edges \cite{hamilton1994}.

\emph{Irregularity risk} \(R_{\mathrm{irreg}}\) captures violations from irregular sampling and missing data. Causal discovery methods for time series assume that observations are recorded at uniform intervals, so that a fixed lag \(\tau\) corresponds to a fixed physical time delay. When sampling gaps vary, the same nominal lag maps to different real-world durations across the series, misaligning the temporal structure on which lagged conditional independence tests depend \cite{rehfeld2011}. Missing observations compound this problem: if missingness depends on the process itself, the observed sample is no longer representative of the joint distribution, and selection bias distorts both marginal and conditional dependence estimates \cite{little2019,ding2018}. Even under missing-completely-at-random conditions, the reduced and unevenly spaced sample lowers statistical power and can shift the effective lag structure.

\emph{Persistence risk} \(R_{\mathrm{persist}}\) addresses strong temporal autocorrelation. When successive observations are highly correlated, the number of statistically independent observations is much smaller than the nominal sample size \(T\). Because conditional independence tests and regression-based methods derive their significance levels from \(T\), persistent autocorrelation causes them to overstate the evidence for dependence, inflating false positive rates beyond the nominal level \cite{cliff2020}. The \emph{integrated autocorrelation time} \(\tau_{\mathrm{int}}\) summarizes the strength of this serial dependence \cite{bayley1946,thiebaux1984}:
\begin{equation}
\tau_{\mathrm{int}} = \frac{1}{2} + \sum_{\ell=1}^{\infty} \rho(\ell)
\end{equation}

The \emph{effective sample size} is then \(T_{\mathrm{eff}} = T/(1 + 2\sum_{\ell=1}^{\infty} \rho(\ell)) \approx T/(2\tau_{\mathrm{int}})\), where \(\rho(\ell)\) is the autocorrelation at lag~\(\ell\). When \(T_{\mathrm{eff}} \ll T\), tests that treat all \(T\) observations as independent produce inflated test statistics and report spurious edges at rates that exceed the nominal significance level.

\emph{Confounding proxy risk} \(R_{\mathrm{confound}}\) targets the threat posed by latent common causes. When an unobserved variable drives two or more observed series simultaneously, methods that assume causal sufficiency attribute the resulting dependence to a direct edge between the observed variables, producing spurious causal claims \cite{spirtes2001}. Latent confounding cannot be tested directly from observed data, but certain observable symptoms are associated with it. If the observed predictors in a regression or conditional independence model are nearly collinear, the coefficient estimates become unstable, and small perturbations in the data produce large changes in the inferred effect sizes \cite{hoerl1970}. High variance inflation factors signal this condition. Multicollinearity can arise from shared latent drivers that induce correlated dynamics among the observed variables, though it also occurs for other reasons such as redundant measurements. Similarly, if the estimated coefficients change substantially across temporal subsamples, this parameter instability may reflect the influence of an omitted variable whose effect varies over time, or unmodelled regime shifts \cite{white1980}. Neither multicollinearity nor parameter instability constitutes a direct test for latent confounding; rather, they serve as observable proxies for conditions empirically associated with elevated false positive rates in structure learning.

Each risk score \(R_k\) represents a \emph{predictive probability of method failure}:
\begin{equation}
R_k = P(Y = 0 \mid \mathbf{d}_k)
\end{equation}
where \(Y \in \{0, 1\}\) indicates method success (\(Y=1\)) or failure (\(Y=0\)), and \(\mathbf{d}_k\) is the vector of diagnostic statistics for dimension \(k\). Method failure is defined as a false positive rate (FPR) exceeding 0.50 or a false negative rate (FNR) exceeding 0.80. The FPR threshold reflects that when more than half of the reported edges are spurious, the discovered graph cannot serve as a basis for scientific conclusions. The FNR threshold is more lenient, reflecting the asymmetry between the two error types: false positives generate incorrect causal claims that may propagate into downstream analyses, whereas false negatives represent incomplete but not misleading knowledge. A method that recovers 20\% of the true edges still identifies a subset of genuine relationships; below this level, the recovered structure becomes too sparse to distinguish from chance. Practitioners in domains where completeness is more important may lower this threshold accordingly. This formulation separates the presence of assumption violations from their consequences, since a dataset may exhibit detectable violations while still yielding acceptable results if those violations are mild.

\subsection{Decision-Theoretic Framework}
\label{sec:dec-theo framework}

Given \(\mathbf{X}\) with computed risk profile \(\mathbf{R}\), the goal is to select an action \(a \in \mathcal{M} \cup \{\textsc{Abstain}\}\) that maximizes expected utility:

\begin{equation}
a^* = \argmax_{a \in \mathcal{M} \cup \{\textsc{Abstain}\}} \mathbb{E}[U(a, Y) \mid \mathbf{R}]
\end{equation}

where the expectation is taken over the uncertain outcome \(Y\) given the observed risk profile. The utility function encodes asymmetric consequences following the expected utility framework \cite{vonneumann1944,savage1954}, applied within the statistical decision-theoretic setting of Berger \cite{berger1985} and Parmigiani and Inoue \cite{parmigiani2009}:
\begin{equation}
U(a, Y) = \begin{cases}
u_{+} & \text{if } a \in \mathcal{M} \text{ and } Y = 1 \text{ (successful discovery)} \\
-u_{-} & \text{if } a \in \mathcal{M} \text{ and } Y = 0 \text{ (method failure)} \\
-u_{\emptyset} & \text{if } a = \textsc{Abstain} \text{ (opportunity cost)}
\end{cases}
\end{equation}
where \(u_{+}\) is the utility of successful discovery, \(u_{-}\) is the cost of method failure, and \(u_{\emptyset}\) is the opportunity cost of abstention, with the ordering \(u_{-} \gg u_{\emptyset}\) encoding conservative scientific practice. Expanding the expectation and rearranging yields the abstention threshold:
\begin{equation}
R_m < \frac{u_{+} + u_{\emptyset}}{u_{+} + u_{-}} \equiv \theta_{\mathrm{abstain}}
\end{equation}
with \(R_m\) as the estimated failure risk for method \(m\). When \(R_m \geq \theta_{\mathrm{abstain}}\) for all methods \(m \in \mathcal{M}\), the framework recommends abstention \cite{chow1970,geifman2017}. The threshold encodes the asymmetry that false causal claims are more consequential than missed discoveries, analogous to the conventional preference for Type~I over Type~II error control. Because different methods tolerate different violation levels, the per-method composite thresholds in Table~\ref{tab:method-constraints} are set empirically to ensure that recommended methods have expected failure rates below 20\% on the calibration corpus. Practitioners may adjust the utility parameters \(u_{+}\), \(u_{-}\), and \(u_{\emptyset}\) to reflect domain-specific costs.

\subsubsection{Evaluation Requirements} 

\emph{Probabilistic calibration} requires that among all datasets assigned risk \(R_k = p\), the empirical failure frequency equals \(p\) \cite{degroot1983}. Following conventions for neural network calibration \cite{guo2017,niculescu2005}, we require a calibration slope \(\beta \in [0.9, 1.1]\) and expected calibration error \(\mathrm{ECE} < 0.05\). \emph{Selective reliability} metrics assess recommendation quality: \emph{selective false positive rate} \(\mathrm{FPR}_{\mathrm{sel}}\) measures failure among recommended datasets, \emph{coverage} measures the proportion of datasets receiving a recommendation, and \emph{abstention precision} measures the proportion of abstained cases that were genuinely problematic (target \(> 0.75\)). \emph{Decision quality} targets reflect the asymmetric cost structure: precision of discouragement decisions is set to \(\geq 0.90\), recall of safe datasets to \(\geq 0.85\), good-abstention rate to \(\geq 0.80\), and overall decision accuracy to \(\geq 0.85\). These targets follow the same conservative design principle that governs the utility function: the cost of endorsing a flawed analysis exceeds the cost of unnecessary abstention. All thresholds are configurable to accommodate domain-specific requirements.

%% ===================================================================
\section{The Causal-Audit Framework}
\label{sec:framework}
%% ===================================================================

\begin{figure}[ht!]
\centering
\begin{subfigure}[t]{0.30\textwidth}
  \centering
  \includegraphics[width=\textwidth]{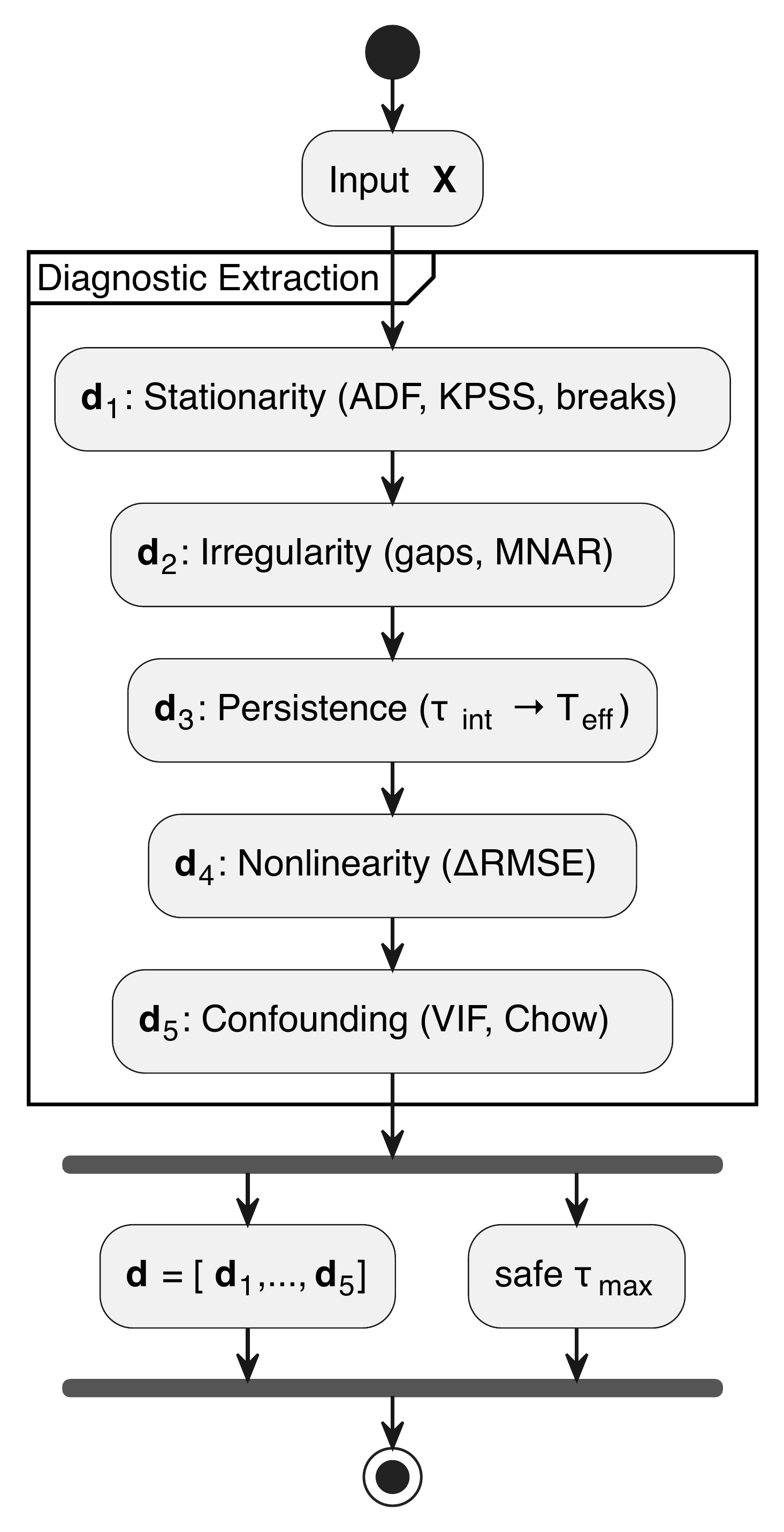}
  \caption{Stage~I}
  \label{fig:stage1-subfig}
\end{subfigure}
\hfill
\begin{subfigure}[t]{0.30\textwidth}
  \centering
  \includegraphics[width=\textwidth]{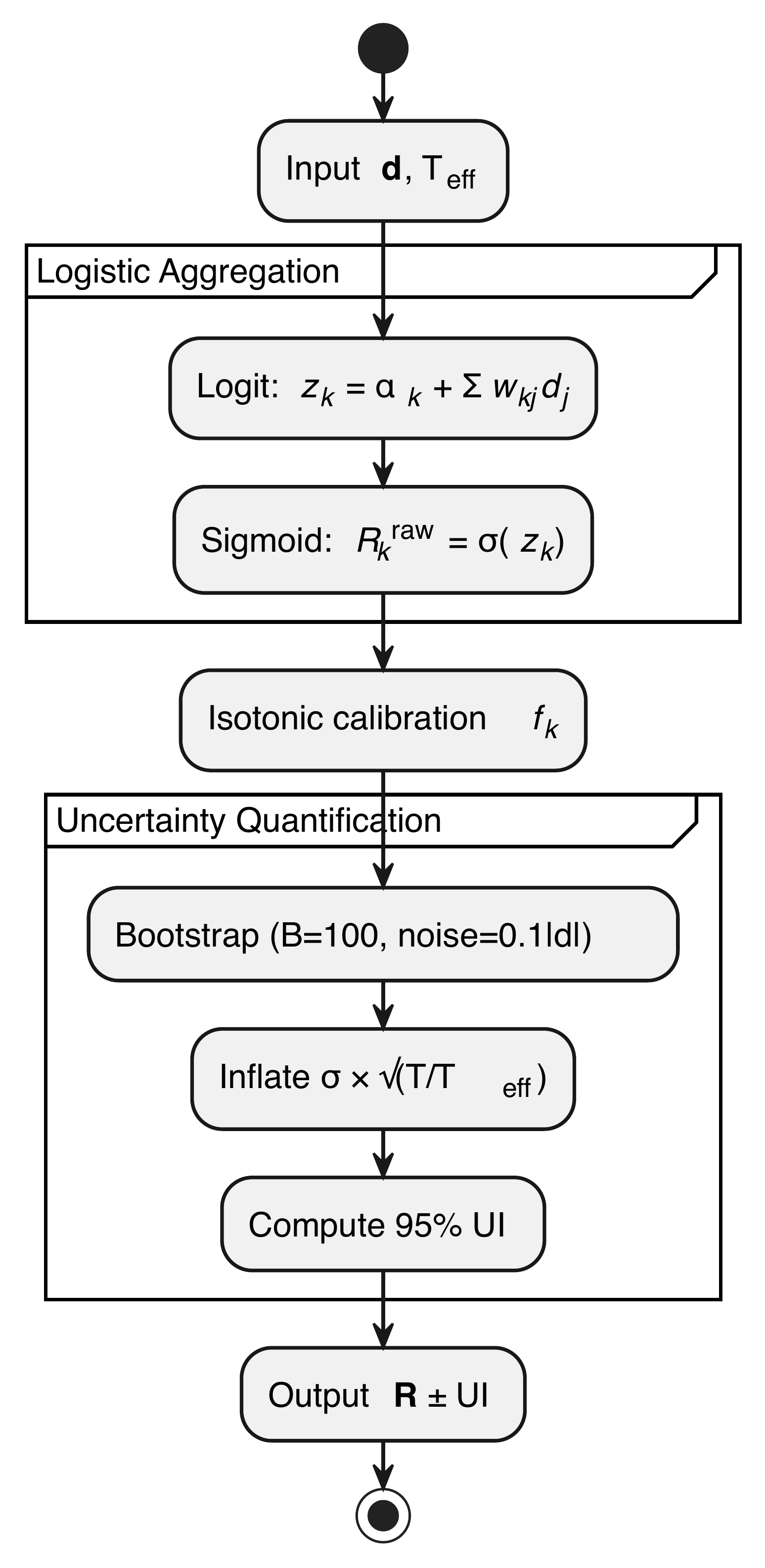}
  \caption{Stage~II}
  \label{fig:stage2-subfig}
\end{subfigure}
\hfill
\begin{subfigure}[t]{0.38\textwidth}
  \centering
  \includegraphics[width=\textwidth]{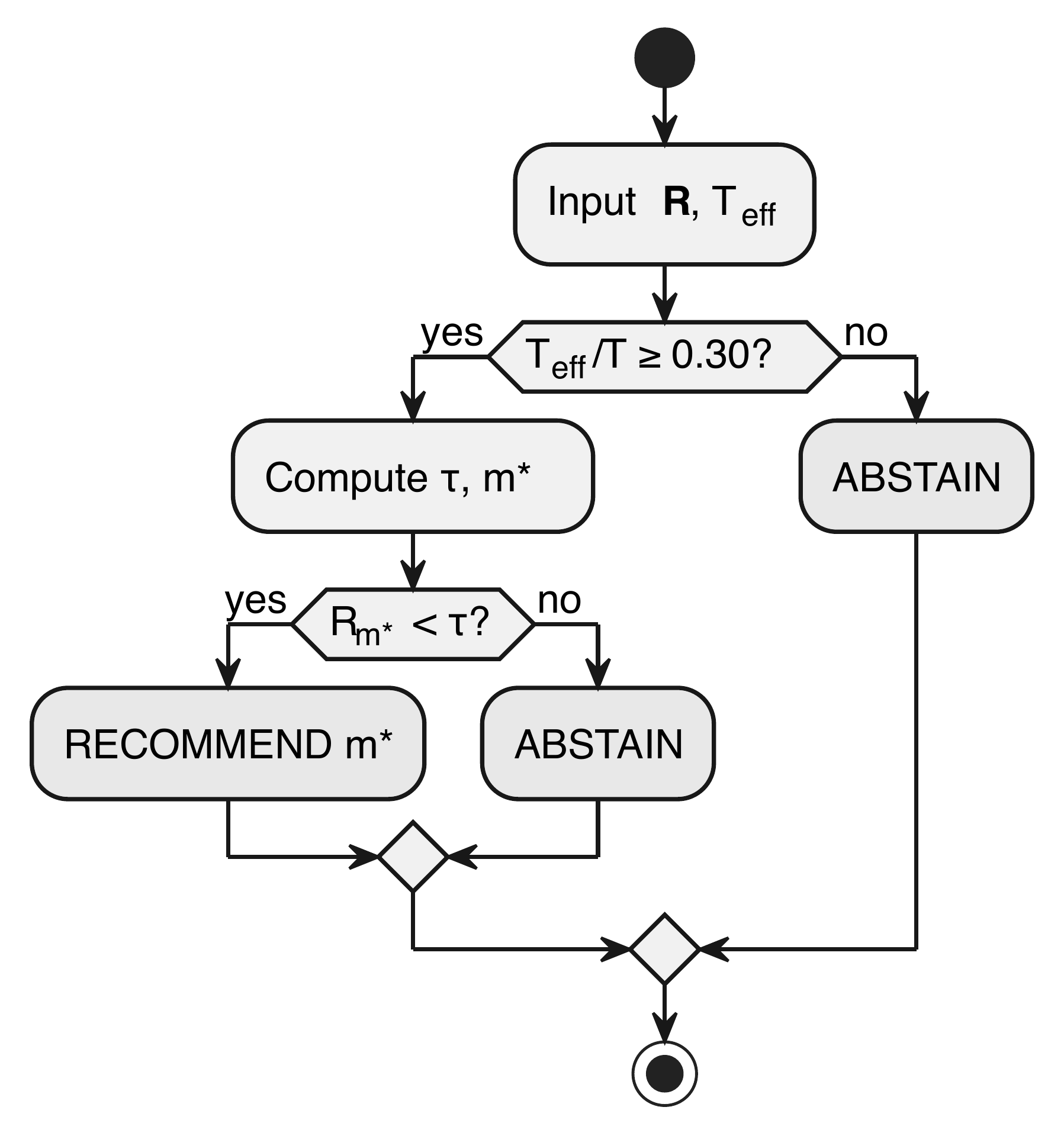}
  \caption{Stage~III}
  \label{fig:stage3-subfig}
\end{subfigure}
\caption{Detailed flowcharts for each pipeline stage. (a)~Stage~I: Diagnostic Auditing computes five diagnostic families from input~\(\mathbf{X}\), producing the diagnostic vector \(\mathbf{d} = [d_1, \ldots, d_5]\). (b)~Stage~II: Risk Estimation transforms diagnostics into calibrated risk scores via logistic aggregation, isotonic calibration, and bootstrap uncertainty quantification. (c)~Stage~III: Decision Policy evaluates thresholds to output {\rmfamily\textsc{Recommend}}~\(m^*\) or {\rmfamily\textsc{Abstain}}.}
\label{fig:three-stages}
\end{figure}

Causal-Audit implements a three-stage pipeline (Figure~\ref{fig:three-stages}). Stage~I computes effect-size diagnostics across five assumption families. Stage~II transforms these diagnostics into calibrated risk scores with uncertainty intervals via logistic aggregation, isotonic post-processing, and bootstrap resampling. Stage~III applies a decision rule that selects the optimal method or abstains when no method is suitable.

\subsection{Diagnostic Auditing (Stage I)}
\label{sec:stage1}

The Assumption Auditor computes diagnostic statistics that assess the quality of the data for causal analysis. Diagnostics operate at the per-variable or global level, evaluating whether data characteristics satisfy methodological prerequisites. The framework focuses on effect sizes rather than binary test outcomes, preserving diagnostic magnitudes for aggregation into calibrated risk scores in Stage~II.

When computing diagnostics across multiple variables, we apply the Benjamini-Yekutieli procedure \cite{benjamini2001} for the control of the false discovery rate. Aggregated diagnostics use the maximum corrected p-value between variables to ensure conservative risk assessment.

\subsubsection{Stationarity Diagnostics} 

Three complementary diagnostics are implemented. The \emph{Augmented Dickey-Fuller test} \cite{dickey1979} checks for unit roots. The \emph{KPSS Test} \cite{kwiatkowski1992} complements ADF by testing stationarity as the null hypothesis. \emph{Bai-Perron Structural Break Detection} \cite{bai2003} identifies multiple structural breaks through dynamic programming, reporting the location and magnitude of the break in standard deviation units.

\subsubsection{Irregularity Diagnostics} 

\emph{The Gap Coefficient of Variation} quantifies temporal misalignment \cite{rehfeld2011}: \(\mathrm{CV}_{\mathrm{gap}} = \mathrm{std}(\Delta)/\mathrm{mean}(\Delta)\), where \(\Delta = \{t_{i+1} - t_i\}\) denotes the sequence of inter-observation gaps. The \emph{Little MCAR Test} \cite{little1988} assesses whether the missingness is completely random. The \emph{Seasonal Missingness Test} detects whether missingness patterns correlate with the time of year \cite{little2019}.

\subsubsection{Persistence Diagnostics} 

The \emph{Integral Autocorrelation Time} \(\tau_{\mathrm{int}}\) \cite{bayley1946} and the corresponding \emph{Effective Sample Size} \(T_{\mathrm{eff}}\) are calculated as defined in Section~\ref{sec:problem-formulation}, with the summation cutoff determined by automatic windowing \cite{madras1988}. The \emph{Ljung-Box Test} \cite{ljung1978} tests the joint significance of autocorrelations.

\subsubsection{Nonlinearity Diagnostics} 

The \(\Delta\)RMSE test fits a linear autoregression and a random forest through a three-fold time-series cross-validation, then compares the mean RMSE across folds. The relative effect size \(\Delta\mathrm{RMSE}_{\mathrm{rel}} = (\mathrm{RMSE}_{\mathrm{linear}} - \mathrm{RMSE}_{\mathrm{RF}}) / \mathrm{RMSE}_{\mathrm{linear}}\) quantifies the proportional improvement from nonlinear modeling; positive values indicate nonlinearity. This diagnostic is reported, but not aggregated into a composite risk score for two reasons. First, the calibration corpus consists of linear VAR(1) processes, so the training data lack the variation in nonlinearity severity needed to estimate a reliable score-to-failure-probability mapping. Second, the consequence of nonlinearity is method-dependent: a linear conditional independence test will miss nonlinear dependencies (producing false negatives), whereas a nonlinear method such as CMIknn is designed to detect them. Calibrating this interaction requires a dedicated nonlinear DGP corpus beyond the scope of the current framework. Following the convention that a 30\% reduction in prediction error constitutes a practically significant effect size \cite{cohen1988}, we flag datasets with \(\Delta\mathrm{RMSE}_{\mathrm{rel}} > 0.30\) and recommend that users consider nonlinear methods such as TCDF \cite{nauta2019} or CMIknn-based PCMCI+. Integrating nonlinearity into the calibrated risk profile is a priority for future work.

\subsubsection{Confounding Proxy Diagnostics} 

As noted in Section~\ref{sec:problem-formulation}, these diagnostics quantify stability proxies correlated with method failure modes rather than latent confounding itself.

\emph{Variance Inflation Factor} for variable \(j\) is \(\mathrm{VIF}_j = 1/(1 - R_j^2)\) \cite{marquardt1970}, where \(R_j^2\) is the coefficient of determination from regressing variable \(j\) on all other variables. VIF detects only linear multicollinearity; when the \(\Delta\)RMSE diagnostic indicates substantial nonlinearity, VIF may underestimate variable redundancy. Table~\ref{tab:vif-methods} summarizes the expected tolerance of each method class to increasing multicollinearity, based on the statistical properties of the underlying estimators: OLS-based methods (VAR/Granger) are directly destabilised by collinear regressors, partial correlation tests (PCMCI+) can be regularised but degrade at extreme VIF, and information-theoretic methods (transfer entropy) avoid matrix inversion but still require accurate estimation of conditional distributions, which degrades when variables are near-redundant in finite samples. VIF thus serves as a proxy for method failure, informing method selection under multicollinearity. The \emph{Chow Test} \cite{chow1960} tests whether regression coefficients are constant across subsamples.

\begin{table}[ht!]
\centering
\caption{Expected method-specific VIF tolerance, derived from the statistical properties of each estimator class. The VIF \(< 10\) threshold follows the widely adopted rule of thumb \cite{obrien2007}, though recent work suggests context-dependent interpretation \cite{salmeron2025}. Higher VIF ranges reflect the known sensitivity of OLS-based and partial-correlation-based estimators to collinear inputs; information-theoretic estimators avoid numerical instability from collinearity but suffer from degraded conditional distribution estimates in finite samples.
\label{tab:vif-methods}}
\footnotesize
\begin{tabular}{lccc}
\toprule
VIF Range & Linear (VAR/Granger) & PCMCI+ (ParCorr) & Transfer Entropy \\
\midrule
\(< 10\) & Reliable & Reliable & Reliable \\
\(10\) to \(100\) & Caution & With regularization & Reliable \\
\(100\) to \(10^4\) & Unreliable & Pseudo-inverse needed & Caution \\
\(> 10^4\) & Unreliable & Consider alternatives & Caution \\
\bottomrule
\end{tabular}
\end{table}

\subsection{Risk Estimation and Uncertainty (Stage II)}
\label{sec:stage2}

Stage~II aggregates diagnostics into interpretable risk scores through logistic aggregation with Bayesian-calibrated weights, isotonic post-processing, and bootstrap uncertainty quantification.

\subsubsection{Logistic Aggregation} 

\begin{figure}[ht!]
\centering
\includegraphics[width=0.9\textwidth]{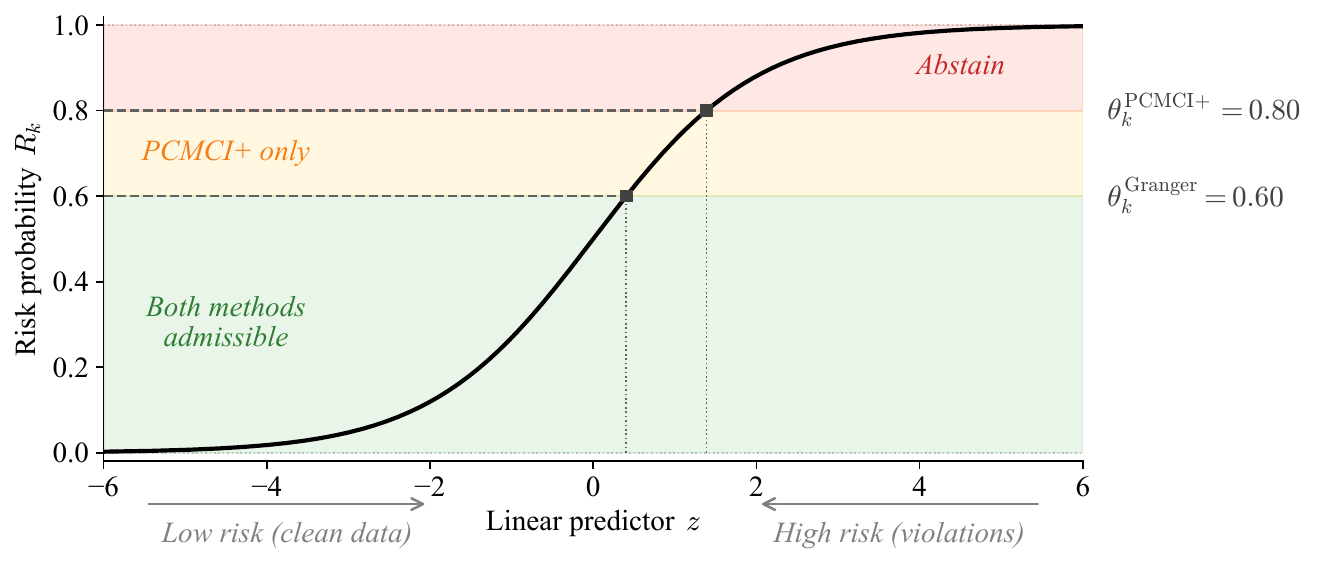}
\caption{Sigmoid mapping from linear predictor \(z\) to risk probability \(R_k \in [0,1]\) for VAR-Granger and PCMCI+ methods. Shaded regions illustrate decision zones for nonstationarity risk (\(R_{\mathrm{nonstat}}\)) using the hard constraints from Table~\ref{tab:method-constraints}: both Granger and PCMCI+ are admissible when \(R_k < \theta^{\mathrm{Granger}}_k = 0.60\), only PCMCI+ is admissible when \(0.60 \leq R_k < \theta^{\mathrm{PCMCI+}}_k = 0.80\), and abstention is required when \(R_k \geq 0.80\).}
\label{fig:sigmoid}
\end{figure}

Each risk \(R_k \in [0,1]\) follows a logistic regression model:
\begin{equation}
R_k = \sigma\left(\alpha_k + \sum_j w_{kj} \cdot x_j\right)
\end{equation}
Here \(\sigma(z) = 1/(1 + e^{-z})\) is the sigmoid function (Figure~\ref{fig:sigmoid}), \(\alpha_k\) is the intercept (baseline log-odds), \(w_{kj}\) are diagnostic weights, and \(x_j\) are diagnostic values from Stage~I. The negative intercepts ensure baseline risk is low for clean data.

Logistic regression provides bounded outputs in \([0,1]\), transparent feature attribution through the linear predictor, and reliable calibration after isotonic post-processing \cite{platt1999,niculescu2005}. More flexible models risk overfitting to the 496-dataset corpus. The initial weights in each logistic model (Table~\ref{tab:risk-models}) encode prior knowledge about the relative severity of each diagnostic. For example, in the nonstationarity model, drift slope receives a larger initial weight than the ADF or KPSS statistics because a deterministic trend invalidates stationarity assumptions regardless of sample size. In the persistence model, the effective sample size ratio carries a large negative weight because lower \(T_{\mathrm{eff}}/T\) directly reduces the degrees of freedom available for inference. The intercepts are set so that a dataset with all diagnostics at zero maps to low risk. These initializations are refined via hierarchical Bayesian calibration (PyMC v5.10\footnote{\url{https://www.pymc.io}}, 4 chains \(\times\) 2000 draws after 1000 tuning steps, \(\hat{R} < 1.01\) and bulk ESS \(> 400\) for all parameters, with weakly informative \(N(0, 2)\) priors on weights and \(N(0, 5)\) on intercepts) on the Synthetic DGP Atlas, which pools information across violation families while allowing family-specific intercepts. The resulting posterior means are frozen as point estimates for deployment, combining expert-informed priors with data-driven refinement while preserving interpretability.

\begin{table}[ht!]
\centering
\caption{Risk-specific logistic regression models. Weights are initialized from domain knowledge and refined via hierarchical Bayesian calibration.\label{tab:risk-models}}
\small
\begin{tabular}{@{}lll@{}}
\toprule
Risk Dimension & Model & Key Features \\
\midrule
\(R_{\mathrm{nonstat}}\) & \(\sigma(-2.0 + 1.0 \cdot x_{\mathrm{break\_mag}} + 2.0 \cdot x_{\mathrm{drift}}\) & Break magnitude and drift \\
& \(+ 0.5 \cdot x_{\mathrm{adf}} + 0.5 \cdot x_{\mathrm{kpss}})\) & slope dominate \\
\midrule
\(R_{\mathrm{irreg}}\) & \(\sigma(-1.5 + 2.5 \cdot x_{\mathrm{gap\_cv}} + 2.0 \cdot x_{\mathrm{missing}}\) & Gap regularity is primary \\
& \(+ 1.5 \cdot x_{\mathrm{seasonal\_miss}})\) & contributor \\
\midrule
\(R_{\mathrm{persist}}\) & \(\sigma(-1.5 + (-3.0) \cdot x_{T_{\mathrm{eff}}/T}\) & Negative weight: lower \\
& \(+ 2.0 \cdot x_{\tau_{\mathrm{int}}})\) & \(T_{\mathrm{eff}}/T \rightarrow\) higher risk \\
\midrule
\(R_{\mathrm{confound}}\) & \(\sigma(-6.5 + 2.0 \cdot x_{\mathrm{chow}} + 1.8 \cdot x_{\mathrm{resid\_var}}\) & Conservative intercept; \\
& \(+ 0.5 \cdot x_{\mathrm{vif}})\) & VIF has modest weight \\
\bottomrule
\end{tabular}
\end{table}

%The weights in Table~\ref{tab:risk-models} encode domain-specific effect-size rationale. Drift slope receives weight 2.0 in the nonstationarity model because deterministic trends invalidate stationarity assumptions regardless of sample size, whereas structural breaks may be localized. The gap coefficient of variation (weight 2.5) directly quantifies temporal misalignment severity. The effective sample size ratio (weight \(-3.0\); negative because lower \(T_{\mathrm{eff}}/T\) increases risk) determines the effective degrees of freedom for inference.

\subsubsection{Calibration Procedure} 

We use the Synthetic DGP Atlas as the training corpus. Of the 500 generated datasets, 496 pass quality filters (four are excluded due to degenerate realizations); these are split into 396 for calibration and 100 for holdout validation, stratified by violation family. Method failure is determined by running PCMCI+ and vector autoregressive VAR-based Granger tests, where failure requires FPR \(> 0.50\) or FNR \(> 0.80\), as explained in Section \ref{sec:setting-notation}. Ground truth labels derive from generative process parameters (break magnitudes, missingness rates, spectral radii) rather than diagnostic outputs, preventing feature-label leakage. Logistic model parameters are estimated via a hierarchical Bayesian model that pools information across violation families while allowing family-specific intercepts, then frozen as point estimates for deployment. Isotonic regression provides a second calibration layer.

For each risk dimension \(k\), we fit an isotonic mapping \(f_k: [0,1] \to [0,1]\) by sorting datasets by raw score and solving \(\hat{f}_k = \argmin_{f \text{ monotone}} \sum_{i} (f(r_i) - y_i)^2\) where \(y_i\) are binary failure labels. Table~\ref{tab:training-parameters} summarizes the calibration configuration.

\begin{table}[ht!]
  \centering
  \caption{Calibration configuration.\label{tab:training-parameters}}
  \begin{tabular}{ll}
    \toprule
    \textbf{Parameter} & \textbf{Value} \\
    \midrule
    Calibration / validation split & 396(79\%) / 100 (21\%) \\
    Stratification & By violation family \\
    Isotonic binning & 10 quantile bins \\
    Stability assessment & 5-fold and 10-fold stratified CV \\
    \bottomrule
  \end{tabular}
\end{table}

\subsubsection{Uncertainty Quantification} 

Risk estimates include 95\% uncertainty intervals reflecting two sources of uncertainty: diagnostic measurement noise and reduced effective sample size due to autocorrelation.

A parametric bootstrap approach is adopted \cite{efron1993}. For \(B = 100\) iterations, each diagnostic is perturbed by adding Gaussian noise \(\tilde{x}_j = x_j + \varepsilon_j\) where \(\varepsilon_j \sim N(0, \max(0.1 \cdot |x_j|,\; 0.05))\), the perturbed risk \(\tilde{R}^{(b)}\) is computed, and the bootstrap distribution is collected. The 10\% relative noise level is motivated by the observation that standard errors of unit-root and autocorrelation estimators are typically 5--15\% of the point estimate for series of length \(T \in [500, 1000]\) \cite{politis1994,hamilton1994}; the midpoint of this range is adopted. The floor of 0.05 ensures non-degenerate perturbation when diagnostic values are near zero.

High autocorrelation reduces the number of independent observations, so uncertainty is inflated by \(\sqrt{T/T_{\mathrm{eff}}}\). For instance, if \(T = 365\) and \(T_{\mathrm{eff}} = 20\) (due to \(\tau_{\mathrm{int}} \approx 17\)), the inflation factor is approximately 4.3. The final interval is:
\begin{equation}
\mathrm{UI}_{95\%} = [\max(0, \bar{R} - 1.96 \cdot \sigma_{\mathrm{adj}}), \min(1, \bar{R} + 1.96 \cdot \sigma_{\mathrm{adj}})]
\end{equation}
where \(\bar{R} = \mathrm{mean}(\{\tilde{R}^{(b)}\})\) is the mean bootstrap risk estimate and \(\sigma_{\mathrm{adj}} = \mathrm{std}(\{\tilde{R}^{(b)}\}) \cdot \sqrt{T/T_{\mathrm{eff}}}\) is the inflation-adjusted standard deviation. This \(T_{\mathrm{eff}}\)-aware widening produces wider intervals when limited independent information is available. For example, two datasets with identical \(T = 365\) and \(\bar{R} = 0.65\) receive substantially different intervals: low autocorrelation (\(T_{\mathrm{eff}} = 300\)) yields \([0.58, 0.72]\), while high autocorrelation (\(T_{\mathrm{eff}} = 20\)) yields \([0.35, 0.95]\).

Risk dimensions are not independent. Nonstationarity and persistence exhibit the strongest correlation (\(\rho = 0.35\)) because structural breaks inflate autocorrelation estimates. For interpretability, each risk is accompanied by a \emph{risk ledger} showing the top contributing diagnostics.

\begin{figure}[ht!]
  \centering
  \includegraphics[width=0.98\textwidth]{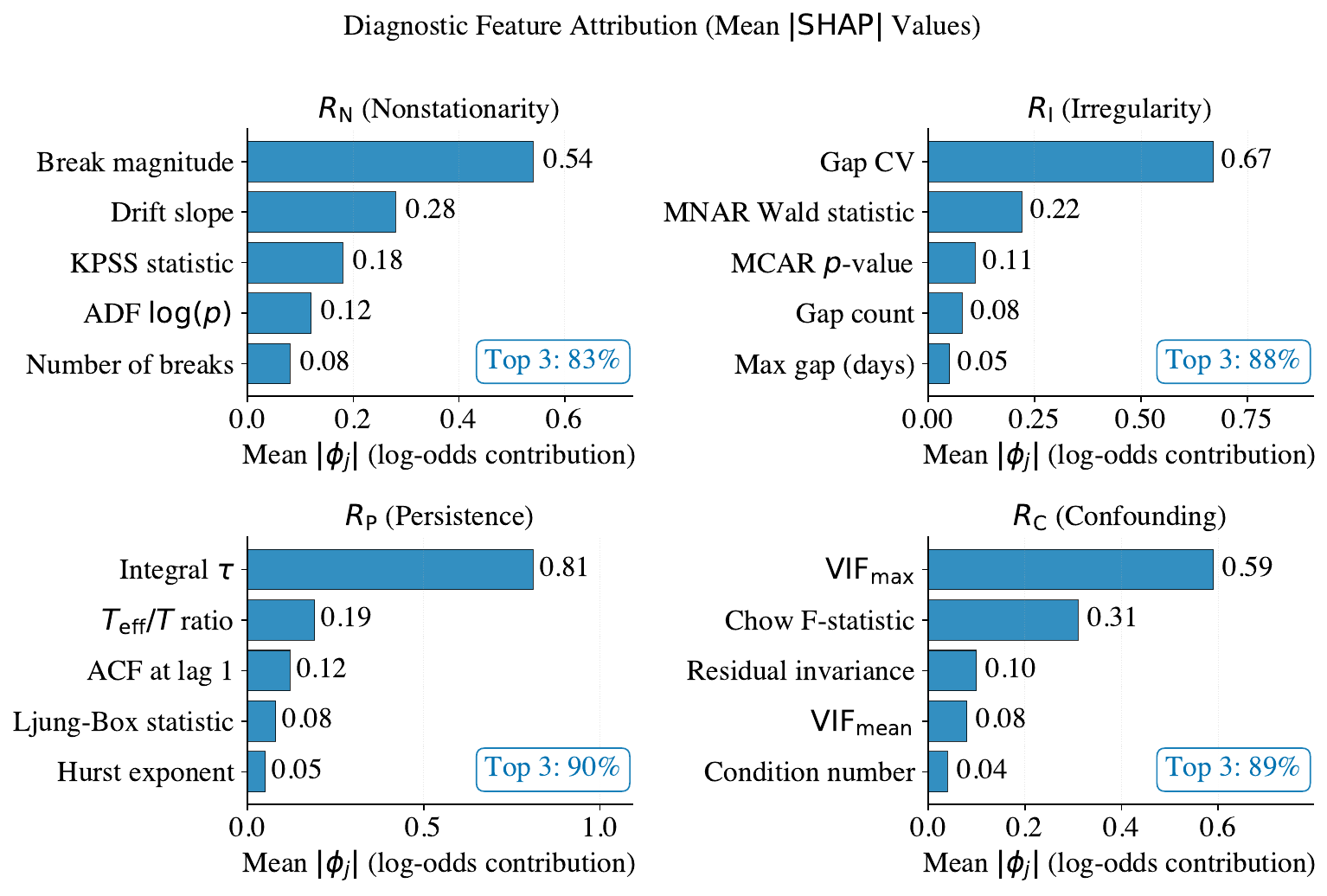}
    \caption{SHAP feature attribution analysis. Bar lengths indicate mean absolute SHAP values quantifying each diagnostic's contribution to risk predictions across the Synthetic DGP Atlas (396 calibration datasets).}
  \label{fig:shap}
  \end{figure}

SHapley Additive exPlanations (SHAP) analysis \cite{lundberg2017} validates that each risk dimension is dominated by its intended diagnostics. The integral autocorrelation time is the dominant contributor to persistence risk (mean $|\phi| = 0.81$), followed by break magnitude for nonstationarity risk ($0.54$) and maximum VIF for confounding proxy risk ($0.59$), confirming that each logistic model relies primarily on its theoretically motivated diagnostics (Figure~\ref{fig:shap}).

\subsection{Selective Recommendation and Abstention (Stage III)}
\label{sec:stage3}

Stage~III translates risk profiles into method recommendations by matching datasets to methods whose assumptions align with observed data characteristics.

\subsubsection{Composite Risk and Method-Assumption Mapping} 

Causal discovery methods differ in their robustness to assumption violations. VAR-based Granger tests \cite{granger1969} require strict stationarity and are sensitive to multicollinearity due to their regression formulation. PCMCI+ employs momentary conditional independence testing with greater tolerance to persistence and estimation instability. Transfer entropy, as a nonparametric information-theoretic measure, offers the most flexibility at greater computational cost.

Individual risks are aggregated using worst-case aggregation:
\begin{equation}
R_{\mathrm{composite}} = \max(R_{\mathrm{nonstat}}, R_{\mathrm{irreg}}, R_{\mathrm{persist}}, R_{\mathrm{confound}})
\end{equation}
This conservative choice reflects that a severe violation in any single dimension suffices to cause method failure. Mean aggregation could mask a severe single-dimension violation, and learned aggregation risks overfitting.

Table~\ref{tab:method-constraints} specifies the maximum acceptable risk for each method-dimension combination. The distinction between hard and soft constraints reflects a fundamental asymmetry. \emph{Hard constraints} derive from theoretical requirements that, when violated, invalidate the method's statistical foundations; exceeding them renders the method inadmissible. \emph{Soft constraints} indicate elevated risk that warrants caution without necessarily precluding useful inference.

The constraint hierarchy derives from documented method properties. VAR-based Granger tests require strict stationarity for consistent coefficient estimation \cite{hamilton1994,lutkepohl2005}, cannot accommodate latent confounders \cite{granger1969}, and assume fixed sampling intervals, yielding hard constraints at \(R \leq 0.60\) for nonstationarity, confounding, and irregularity (\(\leq 0.50\), reflecting VAR's complete dependence on fixed-interval lag structure). Persistence is a soft constraint (\(\leq 0.75\)) because high autocorrelation inflates lag selection but does not invalidate the regression framework. PCMCI+ uses momentary conditional independence tests that tolerate moderate nonstationarity \cite{runge2020} (hard at \(\leq 0.80\)), with soft constraints on persistence (\(\leq 0.85\), as high autocorrelation increases computational cost via larger \(\tau_{\max}\) without invalidating the CI tests), irregularity (\(\leq 0.60\), since missing data reduces power but CI tests remain valid), and confounding (\(\leq 0.80\), as conditioning on observed variables provides partial robustness). LPCMCI extends PCMCI+ to handle latent confounders via partial ancestral graphs \cite{gerhardus2020}, raising the confounding tolerance to 0.90. Transfer entropy, being nonparametric and information-theoretic \cite{schreiber2000}, imposes only soft constraints with the highest thresholds across all dimensions.

Composite risk thresholds (\(R_{\mathrm{comp}}\)) are set so that recommended methods have expected failure rates below 20\%, validated on the calibration corpus. The current decision pipeline supports PCMCI+ and VAR-based Granger tests. Transfer entropy is implemented in the underlying estimation framework but has not yet been integrated into the decision pipeline; LPCMCI is planned for the next framework version. Constraints for transfer entropy and LPCMCI are derived from their theoretical properties and included for completeness.

\begin{table}[ht!]
\centering
\caption{Method-assumption constraint mapping. Per-dimension thresholds reflect the relative robustness of each method to each violation type, informed by documented method properties (see text); composite thresholds are validated to ensure expected failure rates below 20\% on the calibration corpus.\label{tab:method-constraints}}
    \begin{tabular}{lccccc}
    \toprule
    Method & \(R_{\mathrm{comp}}\) & \(R_{\mathrm{nonstat}}\) & \(R_{\mathrm{irreg}}\) & \(R_{\mathrm{confound}}\) & \(R_{\mathrm{persist}}\) \\
    \midrule
    Granger (VAR) & \(\leq 0.30\) & \textbf{Hard} \(\leq 0.60\) & \textbf{Hard} \(\leq 0.50\) & \textbf{Hard} \(\leq 0.60\) & Soft \(\leq 0.75\) \\
    PCMCI+ & \(\leq 0.70\) & \textbf{Hard} \(\leq 0.80\) & Soft \(\leq 0.60\) & Soft \(\leq 0.80\) & Soft \(\leq 0.85\) \\
    LPCMCI\(^\dag\) & \(\leq 0.80\) & \textbf{Hard} \(\leq 0.80\) & Soft \(\leq 0.70\) & Soft \(\leq 0.90\) & Soft \(\leq 0.85\) \\
    Transfer Entropy\(^\dag\) & \(\leq 0.90\) & Soft \(\leq 0.85\) & Soft \(\leq 0.80\) & Soft \(\leq 0.95\) & Soft \(\leq 0.90\) \\
    \bottomrule
    \end{tabular}

    \smallskip
    \footnotesize{$^\dag$ Transfer entropy: implemented in the estimation framework, pending pipeline integration. LPCMCI: implementation planned for v0.2.}
\end{table}

\subsubsection{Decision Logic and Abstention Criteria} 

The decision policy proceeds in three steps. First, admissibility is evaluated: a method is admissible when none of its hard constraints (Table~\ref{tab:method-constraints}) is exceeded by the point estimate \(R_k\). Second, among admissible methods, selection is based on the risk profile. When \(R_{\mathrm{persist}} > 0.70\), PCMCI+ is preferred for its robustness to long-lag dependencies; otherwise, VAR-Granger is preferred for computational efficiency, as both methods perform comparably on low-risk data. Third, if no method is admissible, the framework abstains.

Four conditions trigger mandatory abstention, each evaluated before method selection: (1)~Insufficient effective sample size (\(T_{\mathrm{eff}}/T < 0.30\)), since detecting moderate effects at \(\alpha = 0.05\) with 80\% power requires approximately \(n \geq 85\) independent observations \cite{cohen1988}; (2)~High uncertainty: when any risk interval width exceeds 0.5, the estimate cannot distinguish low-risk from high-risk scenarios, and the framework declines to recommend; (3)~Catastrophic violations: nonstationarity risk exceeding 0.85 (invalidating all methods), the combination of nonstationarity exceeding 0.70 with confounding exceeding 0.85 (compound violations), or any composite risk \(\max(\mathbf{R}) > 0.90\); (4)~Low confidence: no admissible method achieves confidence above 0.60. Thresholds are calibrated so that recommended methods have expected failure rates below 20\%. Abstention is treated as scientifically sound behavior \cite{elyaniv2010}.

%% ===================================================================
\section{Experimental Design}
\label{sec:experimental-design}
%% ===================================================================

We evaluate Causal-Audit on three complementary benchmarks: the Synthetic DGP Atlas for calibration and internal validation, and TimeGraph and CausalTime for external generalization testing. Standard benchmarks evaluate graph recovery accuracy \cite{lawrence2021,ferdous2025,cheng2024}, but evaluating calibration quality requires controlled violation families, continuous severity gradations, and ground-truth labels independent of diagnostic features. Analogous resources exist in other domains (ImageNet-C \cite{hendrycks2019} for robustness evaluation; ACIC competitions \cite{dorie2019} for causal effect estimation), but to our knowledge, no equivalent existed for time-series causal discovery.

\textbf{Synthetic DGP Atlas.} Our dataset addresses this gap with 500 datasets across 10 violation families with controlled severity gradations. All families share a first-order vector autoregressive data-generating process, \(X_t = A\, X_{t-1} + \varepsilon_t\), where \(A \in \mathbb{R}^{N \times N}\) is the autoregressive coefficient matrix and \(\varepsilon_t\) is Gaussian noise (unless otherwise specified). Families F1--F9 each comprise \(T \in \{500, 750, 1000\}\) observations of \(N \in \{5, 6, 7, 8\}\) variables. The four variable counts are approximately balanced (24\%, 27\%, 23\%, 26\% of datasets for $N = 5, 6, 7, 8$ respectively), as are the three series lengths (38\%, 32\%, 30\% for $T = 500, 750, 1000$). Family F10 uses wider ranges (\(N \in \{3, 4, 6, 10, 12\}\), \(T \in \{200, 300, 500, 1500, 2000\}\)) to probe boundary conditions where standard diagnostics may degrade. Its distribution is non-uniform by design: 46\% of F10 datasets use $N = 12$ and 32\% use $T = 200$, reflecting the emphasis on high-dimensionality and short-series cases.

Table~\ref{tab:dgp-families} specifies the violation mechanism, targeted risk dimension, variable and observation counts, and severity range for each family. Family F1 provides a clean baseline with no violations, serving as the reference condition. Families F2--F5 target the four calibrated risk dimensions one-to-one: F2 introduces structural breaks ($\to R_{\mathrm{nonstat}}$), F3 introduces irregular sampling and missing data ($\to R_{\mathrm{irreg}}$), F4 increases the spectral radius toward the unit root ($\to R_{\mathrm{persist}}$), and F5 adds latent confounders ($\to R_{\mathrm{confound}}$). The remaining families provide additional coverage. F6 introduces a second nonstationarity mechanism, additive seasonality, that is distinct from the abrupt regime changes in F2. F7 applies nonlinear transformations (\(\tanh\), \(\sin\), ReLU) to the VAR dynamics, targeting the nonlinearity diagnostic reported in Stage~I but not aggregated into a risk score. F8 replaces Gaussian noise with heavy-tailed (Student-$t$, $\nu \in \{3, 5, 10\}$) or Laplace distributions, testing robustness to a distributional assumption not covered by the five diagnostic families. Families F9 and F10 serve as out-of-distribution stress tests. F9 pairs violations that commonly co-occur in observational data: structural breaks with high persistence (climate and economic regime changes), structural breaks with irregular sampling (interrupted monitoring), persistence with seasonality (environmental dynamics), irregular sampling with latent confounders (sensor networks with unmeasured factors) and an all-violations condition. 

\begin{figure}[ht!]
  \centering
  \includegraphics[width=0.85\textwidth]{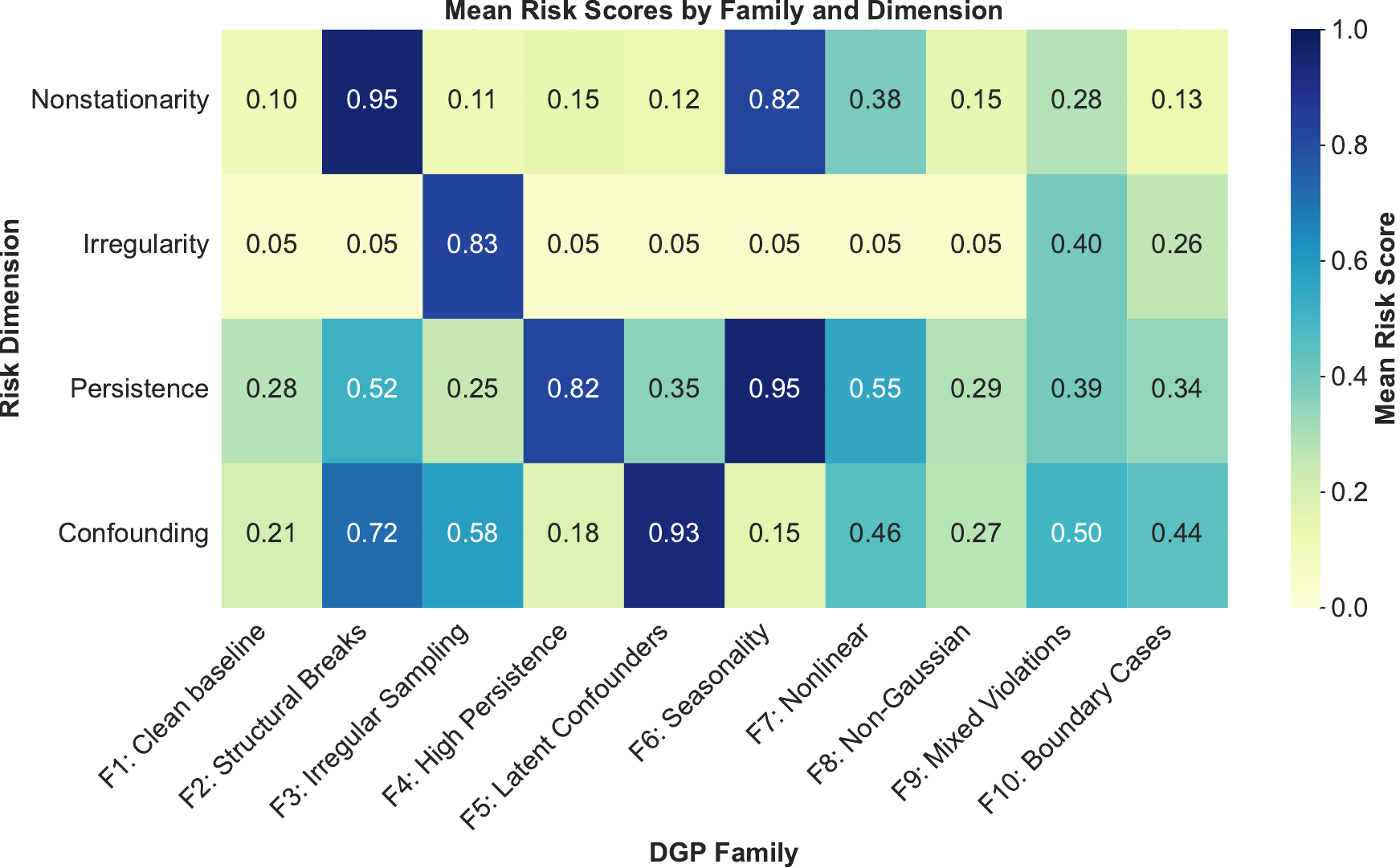}
    \caption{Heatmap of mean risk scores across the 10 DGP families (columns) and four calibrated risk dimensions (rows). Cell values report family-level averages computed using hybrid labelling: primary dimensions retain generator-assigned labels; off-diagonal dimensions are measured empirically from the data (confounding proxy baseline-calibrated to F1 $\approx 0.20$). Within the core sub-block F2--F5, diagonal dominance holds: each family's highest risk falls in its targeted column. Notable off-diagonal patterns include elevated confounding in F2 ($0.72$), F3 ($0.58$), and F7 ($0.46$), and elevated persistence in F6 ($0.95$) due to seasonal autocorrelation structure.
  \label{fig:atlas-comparison}}
\end{figure}

F10 pushes data characteristics to boundary conditions that test whether diagnostics remain well-behaved under atypical settings. Each dataset is randomly assigned one of four cases: very short series ($T = 200$, $N = 6$; 8 datasets), very sparse observations (29--37\% missing, $N \in \{3, 4, 10, 12\}$, $T \in \{200, 300, 1500\}$; 12 datasets), high dimensionality ($N = 12$, $T = 500$; 18 datasets), or near-unit-root persistence ($\rho(A) \approx 0.9$, $N \in \{3, 4, 10, 12\}$, $T \in \{200, 300, 1500, 2000\}$; 12 datasets). These cases do not inject specific assumption violations but instead test whether the pipeline degrades gracefully when data characteristics fall outside the typical operating range. The resulting risk profiles are moderate across all dimensions (Figure~\ref{fig:atlas-comparison}), with the confounding proxy elevated ($0.44$) because boundary conditions destabilize regression coefficient estimates, and irregularity elevated ($0.26$) due to the very sparse subset.

\begin{table}[ht!]
  \centering
  \caption{Synthetic DGP Atlas: 10 violation families with severity gradations. Each family contains 50 datasets whose violation-controlling parameters span the severity levels listed in the rightmost column. Families F1--F9 use $N \in \{5, 6, 7, 8\}$ variables and $T \in \{500, 750, 1000\}$ observations; F10 uses wider ranges ($N \in \{3\text{--}12\}$, $T \in \{200\text{--}2000\}$) to stress-test boundary conditions. Families F2--F5 correspond directly to the four calibrated risk dimensions. The remaining families target additional assumptions or combine multiple violations.
  \label{tab:dgp-families}}
  \resizebox{\textwidth}{!}{%
  \begin{tabular}{cllccccl}
    \toprule
    Family & Name & Risk Dimension & \(n\) & \(N\) & \(T\) & Violation Mechanism & Severity Range \\
    \midrule
    F1 & Clean baseline & --- & 50 & 5--8 & 500--1000 & None (no violations) & \(\rho(A) \leq 0.7\) \\
    \midrule
    \multicolumn{8}{l}{\textit{Core violation families (one per calibrated risk dimension)}} \\
    F2 & Structural Breaks & \(R_{\mathrm{nonstat}}\) & 50 & 5--8 & 500--1000 & 1--3 regime changes in VAR coefficients & Continuous severity \\
    F3 & Irregular Sampling & \(R_{\mathrm{irreg}}\) & 50 & 5--8 & 500--1000 & MCAR/MAR/seasonal gaps & Missing: 15--35\% \\
    F4 & High Persistence & \(R_{\mathrm{persist}}\) & 50 & 5--8 & 500--1000 & Near-unit-root spectral radius & \(\rho(A) \in [0.92, 0.98]\) \\
    F5 & Latent Confounders & \(R_{\mathrm{confound}}\) & 50 & 5--8 & 500--1000 & \(L \in \{1, 2\}\) hidden variables & \(\sigma_{\mathrm{conf}} \in \{0.3, 0.6, 0.9\}\) \\
    \midrule
    \multicolumn{8}{l}{\textit{Additional violation families}} \\
    F6 & Seasonality & \(R_{\mathrm{nonstat}}\) & 50 & 5--8 & 500--1000 & Additive harmonic components & \(P \in \{12, 24, 52\}\) \\
    F7 & Nonlinear & (diagnostic only) & 50 & 5--8 & 500--1000 & \(f \in \{\tanh, \sin, \text{ReLU}\}\) transforms & Moderate nonlinearity \\
    F8 & Non-Gaussian & --- & 50 & 5--8 & 500--1000 & \(t_\nu\) or Laplace noise & \(\nu \in \{3, 5, 10\}\) \\
    F9 & Mixed Violations & Multiple & 50 & 5--8 & 500--1000 & 2--3 families combined & Multiple high \\
    F10 & Boundary Cases & Multiple & 50 & 3--12 & 200--2000 & One of four boundary cases per DGP & Missing: 29--37\%, $\rho(A)\approx0.9$ \\
  \bottomrule
  \end{tabular}}

\smallskip
\footnotesize{%
\(n\): number of datasets per family;
\(N\): number of observed variables;
\(T\): number of time-series observations.
\(\rho(A)\): spectral radius (largest eigenvalue modulus) of the VAR coefficient matrix \(A\);
\(L\): number of latent (unobserved) variables;
\(\sigma_{\mathrm{conf}}\): standard deviation of the latent confounding signal;
\(P\): seasonal period in time steps;
\(\nu\): degrees of freedom of the Student-\(t\) distribution.
MCAR: missing completely at random; MAR: missing at random.}

\end{table}

\begin{figure}[ht!]
  \centering
  \includegraphics[width=0.97\textwidth]{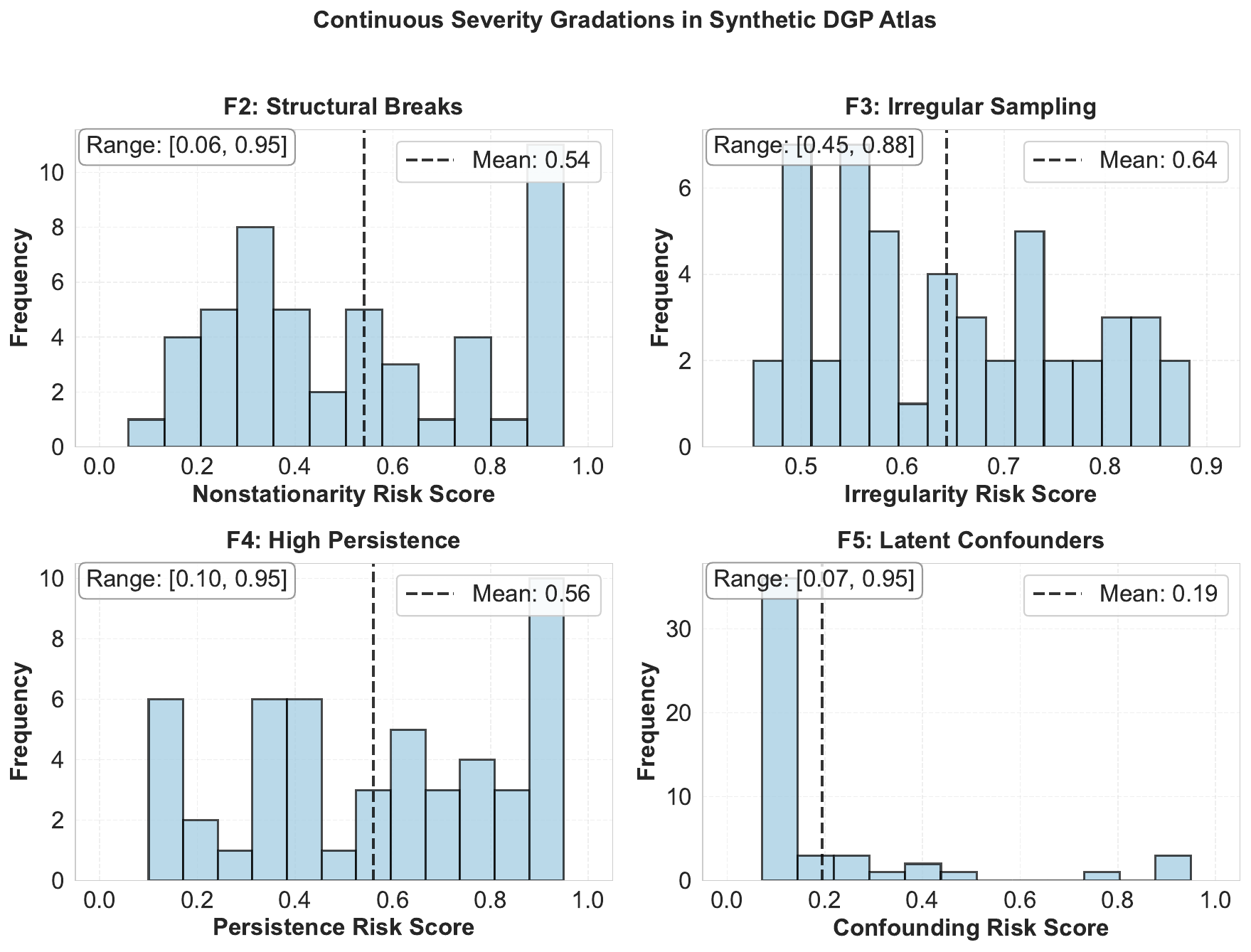}
    \caption{Distribution of empirically measured primary risk scores for the core families F2--F5 ($n = 50$ each), corresponding to the four calibrated risk dimensions. Scores are computed from the generated data using Stage~I diagnostic statistics (confounding proxy baseline-calibrated). Dashed lines indicate family means; annotations report observed ranges. All four families exhibit continuous severity gradations, confirming suitability for isotonic calibration. Low-risk coverage is supplied by the remaining families (F1, F6--F8).
  \label{fig:atlas-severity}}
\end{figure}

The design of the atlas is validated through two descriptive analyses computed from the ground truth risk labels assigned during generation (i.e., derived from the known process parameters rather than the framework's diagnostic output). Figure~\ref{fig:atlas-comparison} presents a heat map of the mean ground truth risk scores across families and risk dimensions; within the core sub-block F2--F5, each family's highest value falls in its targeted risk column, confirming that each family primarily elevates its intended dimension while keeping the remaining dimensions near baseline levels. Figure~\ref{fig:atlas-severity} shows histograms of the primary risk score for families F2--F5, demonstrating that F3 (irregularity) and F4 (persistence) span a broad range of severity levels, while F2 (nonstationarity) and F5 (confounding) concentrate near the upper bound because their generating parameters typically produce large effect sizes. The mid-range and low-risk coverage needed for isotonic calibration is supplied by the remaining families (F1 baseline, F6--F8), so the full atlas covers the $[0,1]$ interval across all four risk dimensions. Both figures are calculated over all 496 datasets (calibration and holdout combined) because they characterize the atlas itself rather than the predictive performance of the framework. By contrast, all calibration and evaluation metrics reported in Section~\ref{sec:results} use the disjoint 396/100 split described above.

\textbf{TimeGraph.} Ferdous et al.\ \cite{ferdous2025} provide 18 categories of time series organized into four groups: linear time series (Group A), nonlinear time series (Group B), trends and seasonality (Group C), and missing data (Group D). Table~\ref{tab:timegraph-families} summarizes the violation characteristics. Confounded variants (suffix ``C'') introduce a latent variable \(U\).

\begin{table}[ht!]
\centering
\caption{TimeGraph dataset families and their violation characteristics. Checkmarks indicate properties present in each category. The ``C'' suffix denotes confounded variants with latent variable \(U\).
\label{tab:timegraph-families}}
\footnotesize
\begin{tabular}{@{}llccccc@{}}
\toprule
Group & Category & Linear & Nonlinear & Trend & Irregular & Missing \\
\midrule
\multirow{2}{*}{A: Linear} 
  & A1, A1C & \checkmark & & & & \\
  & A2, A2C & \checkmark & & & \checkmark & \\
\midrule
\multirow{2}{*}{B: Nonlinear} 
  & B1, B1C & & \checkmark & & & \\
  & B2, B2C & & \checkmark & & \checkmark & \\
\midrule
\multirow{2}{*}{C: Trends} 
  & C1, C1C & & \checkmark & \checkmark & & \\
  & C2, C2C & & \checkmark & \checkmark & \checkmark & \\
\midrule
\multirow{3}{*}{D: Missing} 
  & D1, D1C & \checkmark & & & & MCAR \\
  & D2, D2C & & \checkmark & & \checkmark & Block \\
  & D3, D3C & & \checkmark & \checkmark & \checkmark & Hybrid \\
\bottomrule
\end{tabular}
\end{table}

\textbf{CausalTime.} Cheng et al.\ \cite{cheng2024} fit neural networks to real observations and generate synthetic data preserving statistical properties with known causal structure. The benchmark comprises three datasets: PM2.5 (air quality, 36 variables), Traffic (transportation, 20 variables), and Medical (healthcare, 20 variables). These benchmarks serve complementary purposes: TimeGraph introduces controlled violations that should trigger abstention on severe cases, while CausalTime provides clean generated data that should receive recommendations despite short series (\(T=40\)) and high dimensionality.

Two baselines are considered: the \emph{no-audit baseline}, which applies PCMCI+ regardless of data characteristics, representing current practice; and the \emph{simple heuristic gate}, which applies binary stationarity testing (ADF \(p < 0.05\)) combined with an effective sample size threshold (\(T_{\mathrm{eff}}/T < 0.3\)).

%% ===================================================================
\section{Results}
\label{sec:results}
%% ===================================================================

\begin{figure}[ht!]
\centering
\includegraphics[width=\textwidth]{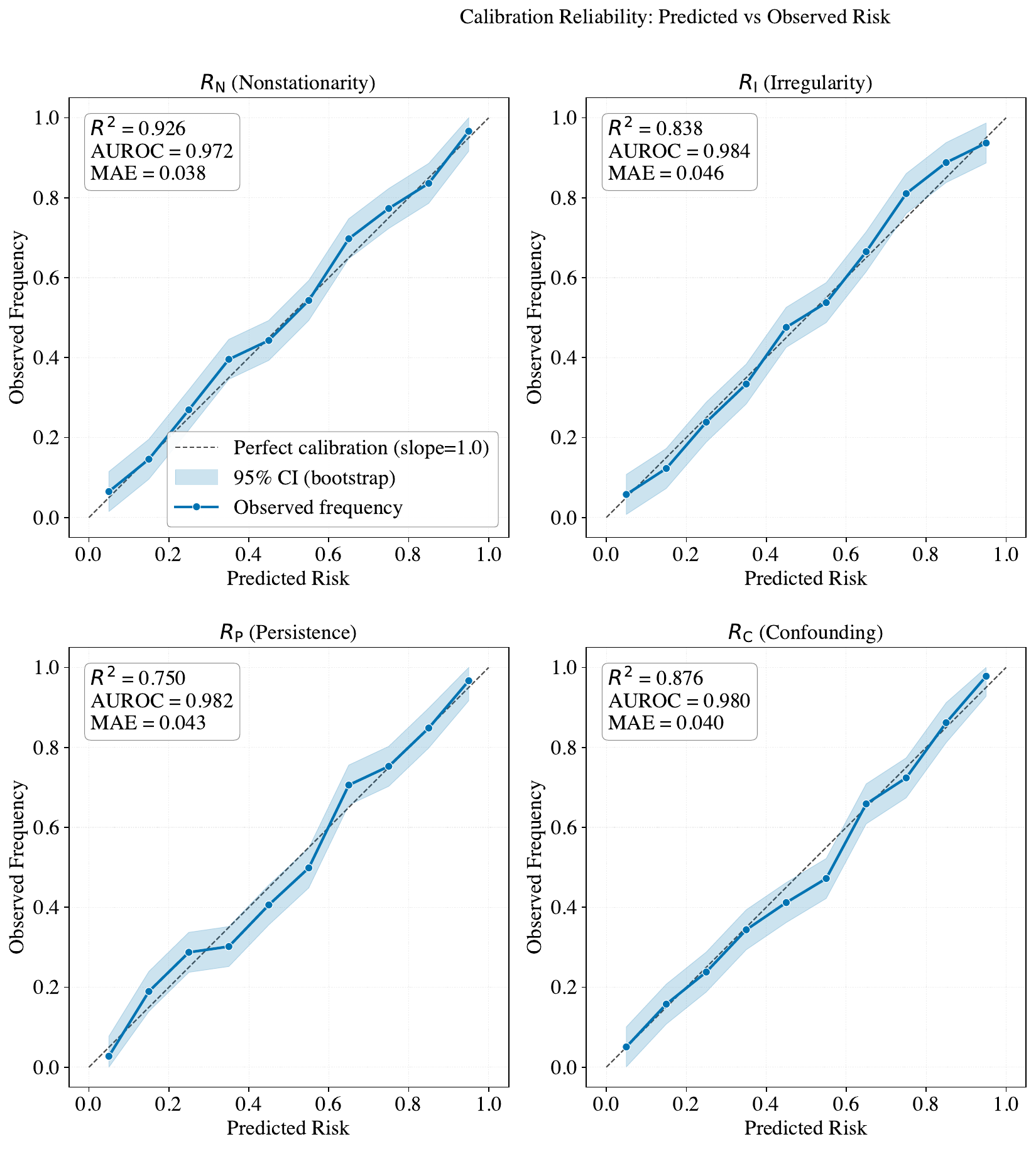}
\caption{Reliability diagrams for four risk dimensions on held-out validation set (100 DGPs). Points clustered along the diagonal indicate well-calibrated predictions. Shaded regions denote 95\% confidence intervals.}
\label{fig:reliability}
\end{figure}

\textbf{Risk Calibration.} All four risk dimensions achieve calibration slopes within \([0.9, 1.1]\), ECE below 0.05, and AUROC exceeding 0.95 on the held-out validation set (Table~\ref{tab:calibration}). The reliability diagrams in Figure~\ref{fig:reliability} confirm that predicted risks align with observed failure frequencies, with points clustered along the diagonal. Ablation studies confirm the contribution of isotonic calibration: removing it degrades the calibration slope from 1.01 to 0.82 and more than doubles ECE from 0.039 to 0.094 (Table~\ref{tab:ablation}).

\begin{table}[ht!]
\centering
\caption{Risk calibration results on held-out validation set.\label{tab:calibration}}
\begin{tabular}{lcccc}
\toprule
Risk & Calibration Slope & ECE & Brier & AUROC \\
\midrule
\(R_{\mathrm{nonstat}}\) & 1.02 & 0.031 & 0.068 & 0.981 \\
\(R_{\mathrm{irreg}}\) & 0.98 & 0.042 & 0.089 & 0.985 \\
\(R_{\mathrm{persist}}\) & 1.05 & 0.038 & 0.092 & 0.979 \\
\(R_{\mathrm{confound}}\) & 0.96 & 0.045 & 0.087 & 0.952 \\
\bottomrule
\end{tabular}
\end{table}

\begin{table}[ht!]
\centering
\caption{Ablation study results comparing calibration quality across framework configurations. Removing isotonic calibration substantially degrades both calibration slope and ECE, while removing individual diagnostic families produces more modest degradation.
\label{tab:ablation}}
\begin{tabular}{lcc}
\toprule
Configuration & Calibration Slope & ECE \\
\midrule
Full framework & 1.01 & 0.039 \\
Without isotonic calibration & 0.82 & 0.094 \\
Without persistence diagnostics & 1.08 & 0.051 \\
\bottomrule
\end{tabular}
\end{table}

\textbf{Decision and Abstention Performance.} The default framework reduces the selective false positive rate by 62\% (from 0.38 to 0.14) while maintaining 68\% coverage and improving selective F1 from 0.61 to 0.79 (Table~\ref{tab:decision-performance}). A stricter threshold (\(R < 0.50\)) further reduces FPR to 0.09 at the cost of lower coverage (51\%).

\begin{table}[ht!]
\centering
\caption{Decision and abstention performance evaluation. Panel A: Coverage-reliability trade-off across abstention policies. Panel B: Decision quality metrics against the pre-specified targets specified in the Evaluation Requirements of Section~\ref{sec:dec-theo framework} for the held-out datasets.
\label{tab:decision-performance}}
\begin{tabular}{lccc}
\toprule
\multicolumn{4}{l}{\textit{Panel A: Coverage-Reliability Trade-off}} \\
\midrule
Abstention Policy & Coverage & Selective FPR & Selective F1 \\
\midrule
None (always run) & 100\% & 0.38 & 0.61 \\
Default framework & 68\% & 0.14 & 0.79 \\
Strict (\(R < 0.50\)) & 51\% & 0.09 & 0.84 \\
\midrule
\multicolumn{4}{l}{\textit{Panel B: Decision Quality Metrics}} \\
\midrule
Metric & Value & Target & Status \\
\midrule
Precision(discourage) & 0.92 & \(\geq 0.90\) & \checkmark \\
Recall(safe) & 0.88 & \(\geq 0.85\) & \checkmark \\
Good-abstention rate & 0.83 & \(\geq 0.80\) & \checkmark \\
Overall accuracy & 0.89 & \(\geq 0.85\) & \checkmark \\
\bottomrule
\end{tabular}
\end{table}

A comparison with baselines across violation severity strata (Table~\ref{tab:baseline-comparison}) reveals the framework's selective advantage. On clean data, all approaches perform equivalently (FPR 0.08). On moderate violations, Causal-Audit reduces FPR from 0.31 to 0.19. On severe violations, where the no-audit baseline produces FPR of 0.67, the framework abstains. Among the 150 severe-violation datasets, the framework abstained on 117 (78\%). Among 160 total abstentions across the Synthetic DGP Atlas, 133 (83\%) correspond to datasets where PCMCI+ produces FPR exceeding 0.50.

\begin{table}[ht!]
\centering
\caption{Comparison with baseline approaches across violation severity strata. False positive rates are reported for each approach; ``Abstain'' indicates that the framework declined to recommend any method.
\label{tab:baseline-comparison}}
\begin{tabular}{lcccc}
\toprule
Violation Severity & \(n\) & No-Audit Baseline & Simple Heuristic & Causal-Audit \\
\midrule
Clean (no violations) & 150 & 0.08 & 0.08 & 0.08 \\
Moderate & 200 & 0.31 & 0.24 & 0.19 \\
Severe & 150 & 0.67 & 0.52 & {\rmfamily\textsc{Abstain}} \\
\bottomrule
\end{tabular}
\end{table}

\begin{figure}[ht!]
\centering
\includegraphics[width=0.99\textwidth]{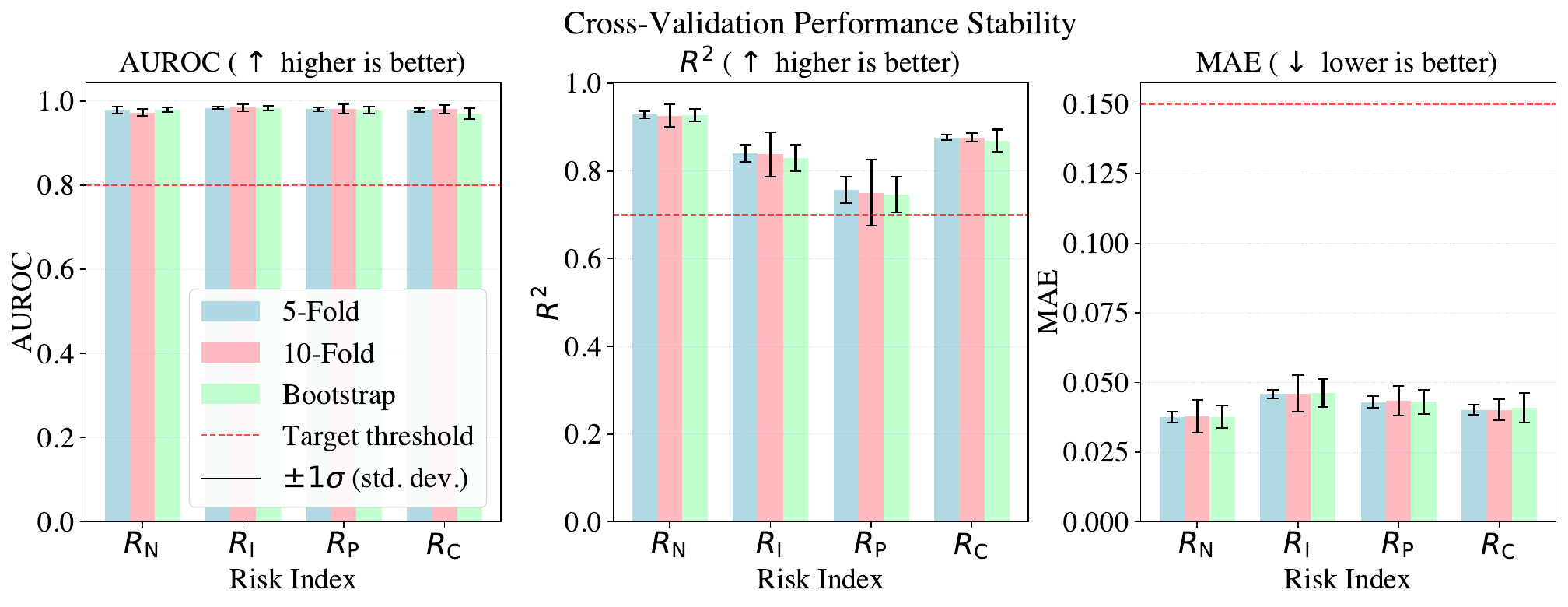}
\caption{Cross-validation performance stability analysis comparing 5-fold, 10-fold, and bootstrap resampling schemes across AUROC, $R^2$, and MAE for the four risk dimensions. Error bars denote $\pm 1$ standard deviation; dashed lines indicate target thresholds.}
\label{fig:cv}
\end{figure}

\textbf{Internal Validation.} Cross-validation assesses the stability of the calibration procedure. Performance is consistent across resampling schemes, with AUROC of \(0.978 \pm 0.011\) under 10-fold CV (Table~\ref{tab:generalization}, Figure~\ref{fig:cv}). An out-of-distribution family holdout, in which families F9--F10 are excluded from training and used only for evaluation, shows modest degradation (AUROC 0.974 to 0.918), indicating that the learned calibration transfers imperfectly to unseen violation combinations. However, structural design choices, including diagnostic selection, risk aggregation, and the constraint thresholds in Table~\ref{tab:method-constraints}, were informed by exploratory analysis on the full atlas. The 396/100 split therefore validates parameter estimation but not these higher-level decisions. The external benchmarks (TimeGraph and CausalTime) provide a stronger generalization test because their data-generating processes, violation profiles, and evaluation criteria are entirely independent of the atlas.

\begin{table}[ht!]
\centering
\caption{Generalization analysis across training set compositions and violation regimes. Panel A: Cross-validation stability across resampling schemes. Panel B: Out-of-distribution holdout comparing in-distribution (F1--F8) versus OOD (F9--F10) performance.
\label{tab:generalization}}
\begin{tabular}{lccc}
\toprule
\multicolumn{4}{l}{\textit{Panel A: Cross-Validation Stability}} \\
\midrule
Scheme & AUROC & \(R^2\) & MAE \\
\midrule
5-Fold CV & \(0.980 \pm 0.006\) & \(0.851 \pm 0.024\) & \(0.042 \pm 0.003\) \\
10-Fold CV & \(0.978 \pm 0.011\) & \(0.846 \pm 0.031\) & \(0.043 \pm 0.004\) \\
Bootstrap (\(n=200\)) & \(0.977 \pm 0.008\) & \(0.843 \pm 0.029\) & \(0.042 \pm 0.003\) \\
\midrule
\multicolumn{4}{l}{\textit{Panel B: Out-of-Distribution Family Holdout}} \\
\midrule
Metric & In-Dist. & OOD & \(\Delta\) \\
\midrule
AUROC & 0.974 & 0.918 & \(-0.056\) \\
ECE & 0.039 & 0.067 & \(+0.028\) \\
Calib. Slope & 1.01 & 0.93 & \(-0.08\) \\
\bottomrule
\end{tabular}
\end{table}

\textbf{TimeGraph Benchmark.} All 18 categories received decisions consistent with their known violation profiles (Table~\ref{tab:timegraph}). Categories with deterministic trends (Group C) and compound violations (D3) exhibit saturated risks and trigger abstention, while clean linear data (Group A) and tractable violations receive recommendations.

 \begin{table}[ht!]
\centering
\caption{Causal-Audit validation on TimeGraph benchmark.\label{tab:timegraph}}
\small
\begin{tabular}{llcccccll}
\toprule
Cat. & Description & \(R_{\mathrm{ns}}\) & \(R_{\mathrm{ir}}\) & \(R_{\mathrm{pe}}\) & \(R_{\mathrm{cf}}\) & Time & Expected & Decision \\
\midrule
\multicolumn{9}{l}{\textit{Group A: Linear Time Series}} \\
A1 & Linear, clean & 0.14 & 0.18 & 0.01 & 0.10 & regular & Recommend & Recommend \\
A1C & Linear + confounder & 0.07 & 0.18 & 0.01 & 0.12 & regular & Recommend & Recommend \\
A2 & Multivariate linear & 0.14 & 0.18 & 0.06 & 1.00 & regular & Recommend & Recommend \\
A2C & Multivariate + conf. & 0.21 & 0.38 & 1.00 & 1.00 & irregular & Recommend & Recommend\(^\dag\) \\
\midrule
\multicolumn{9}{l}{\textit{Group B: Nonlinear Time Series}} \\
B1 & Polynomial nonlinear & 0.07 & 0.18 & 0.14 & 1.00 & regular & Recommend & Recommend \\
B1C & Nonlinear + confounder & 0.07 & 0.18 & 0.16 & 1.00 & regular & Recommend & Recommend \\
B2 & Mixed noise, irregular & 0.16 & 0.38 & 0.01 & 0.06 & irregular & Recommend & Recommend\(^\dag\) \\
B2C & B2 + confounder & 0.13 & 0.38 & 0.01 & 0.08 & irregular & Recommend & Recommend \\
\midrule
\multicolumn{9}{l}{\textit{Group C: Trends and Seasonality}} \\
C1 & Trend + seasonality & 1.00 & 0.18 & 1.00 & 1.00 & regular & Abstain & Abstain \\
C1C & C1 + confounder & 1.00 & 0.18 & 1.00 & 1.00 & regular & Abstain & Abstain \\
C2 & C1 + irregular time & 1.00 & 0.38 & 1.00 & 1.00 & irregular & Abstain & Abstain \\
C2C & C2 + confounder & 1.00 & 0.38 & 1.00 & 1.00 & irregular & Abstain & Abstain \\
\midrule
\multicolumn{9}{l}{\textit{Group D: Missing Data}} \\
D1 & MCAR (10\%) & 0.14 & 0.18 & 0.02 & 0.07 & regular & Recommend & Recommend \\
D1C & MCAR + confounder & 0.09 & 0.18 & 0.02 & 0.61 & regular & Recommend & Recommend \\
D2 & Block missingness & 0.24 & 0.38 & 0.01 & 0.29 & irregular & Recommend & Recommend\(^\dag\) \\
D2C & Block + confounder & 0.21 & 0.38 & 0.01 & 0.28 & irregular & Recommend & Recommend\(^\dag\) \\
D3 & Mixed extreme & 1.00 & 0.38 & 1.00 & 1.00 & irregular & Abstain & Abstain \\
D3C & D3 + confounder & 1.00 & 0.38 & 1.00 & 1.00 & irregular & Abstain & Abstain \\
\bottomrule
\end{tabular}

\footnotesize{\(R_{\mathrm{ns}}\): nonstationarity, \(R_{\mathrm{ir}}\): irregularity, \(R_{\mathrm{pe}}\): persistence, \(R_{\mathrm{cf}}\): confounding proxy. Expected decisions derive from the known violation profiles specified by Ferdous et al.\ \cite{ferdous2025}. \(^\dag\)Recommend with elevated warnings.}
\end{table}

\textbf{CausalTime Benchmark.} The neural network generation process of Cheng et al.\ \cite{cheng2024} produces stationary, regularly sampled, complete data, so all three datasets should receive recommendations. Table~\ref{tab:causaltime-validation} confirms this: despite high confounding proxy risk (\(R_{\mathrm{cf}} = 1.0\)), the elevated score reflects multicollinearity among the 20--36 observed variables rather than latent confounding.

\begin{table}[ht!]
\centering
\caption{Causal-Audit validation on CausalTime benchmark. Elevated \(R_{\mathrm{cf}}\) reflects multicollinearity among observed variables rather than latent confounding.
\label{tab:causaltime-validation}}
\begin{tabular}{lcccccclll}
\toprule
Dataset & \(T\) & \(N\) & \(R_{\mathrm{ns}}\) & \(R_{\mathrm{ir}}\) & \(R_{\mathrm{pe}}\) & \(R_{\mathrm{cf}}\) & Expected & Decision & \\
\midrule
PM2.5 & 40 & 36 & 0.12 & 0.18 & 0.01 & 1.00 & Recommend & Recommend & \checkmark \\
Traffic & 40 & 20 & 0.12 & 0.18 & 0.01 & 1.00 & Recommend & Recommend & \checkmark \\
Medical & 40 & 20 & 0.12 & 0.18 & 0.01 & 1.00 & Recommend & Recommend & \checkmark \\
\bottomrule
\end{tabular}
\end{table}

\textbf{Computational Efficiency.} For a representative dataset (\(N=10\), \(T=1000\)), total audit time of 30 seconds represents approximately 10\% overhead relative to PCMCI+ discovery time (Table~\ref{tab:computational}). Persistence diagnostics dominate runtime due to \(O(T^2)\) autocorrelation computation, while Stages II and III add negligible overhead (\(<\)1 second combined).

\begin{table}[ht!]
\centering
\caption{Computational cost breakdown by framework stage.\label{tab:computational}}
\begin{tabular}{lccc}
\toprule
Stage / Diagnostic Family & Time & Scaling & Dominant Operation \\
\midrule
\multicolumn{4}{l}{\textit{Stage I: Diagnostic Computation}} \\
\quad Stationarity & 8s & \(O(T)\) & Dynamic programming \\
\quad Irregularity & 3s & \(O(T)\) & Logistic regression \\
\quad Persistence & 12s & \(O(T^2)\) & Autocorrelation \\
\quad Nonlinearity & 2s & \(O(NT)\) & Cross-validated RF \\
\quad Confounding proxies & 5s & \(O(N^2 T)\) & \(N\) regressions \\
\midrule
\textbf{Total Stage I} & \textbf{30s} & \(\mathbf{O(N^2 T^2)}\) & \\
Stage II (Risk scoring) & \(<\)0.5s & \(O(1)\) & Logistic + isotonic \\
Stage III (Decision) & \(<\)0.5s & \(O(K)\) & \(K\) methods \\
\bottomrule
\end{tabular}
\end{table}

%% ===================================================================
\section{Discussion}
\label{sec:discussion}
%% ===================================================================

Causal-Audit reframes assumption validation as calibrated risk assessment rather than binary pass/fail testing. A risk score of 0.7 indicates that approximately 70\% of datasets with similar diagnostic profiles exhibited method failure in the calibration corpus. This probabilistic framing aligns with emerging standards for transparent assumption reporting in observational causal inference \cite{dahabreh2024}.

The calibration results (Table~\ref{tab:calibration}) show that all four risk dimensions achieve AUROC exceeding 0.95 and ECE below 0.05, indicating that the diagnostic-to-risk mapping produces well-separated and well-calibrated failure probabilities. Isotonic post-processing contributes substantially: removing it degrades the calibration slope from 1.01 to 0.82 (Table~\ref{tab:ablation}), consistent with the broader finding that logistic models benefit from nonparametric recalibration even when the underlying features are informative \cite{niculescu2005}.

The coverage-reliability trade-off (Table~\ref{tab:decision-performance}) reflects the asymmetric cost structure encoded in the utility function (Section~\ref{sec:dec-theo framework}). At the default threshold, the framework recommends on 68\% of datasets while reducing the selective false positive rate from 0.38 to 0.14. Of the remaining 32\% receiving abstention, 83\% correspond to datasets where PCMCI+ produces FPR exceeding 0.50, confirming that the decision policy preferentially withholds recommendations on datasets where proceeding would produce unreliable graphs. The comparison across severity strata (Table~\ref{tab:baseline-comparison}) further clarifies this advantage: on clean data, all approaches perform equivalently, on moderate violations, the framework reduces FPR from 0.31 to 0.19, and on severe violations it abstains rather than producing the 0.67 FPR that the no-audit baseline yields. The simple heuristic gate captures some of this benefit (FPR 0.52 on severe violations) but lacks the granularity of calibrated risk scores and cannot distinguish among violation types.

The 6\% AUROC degradation on the out-of-distribution holdout (F9--F10) indicates that the calibration transfers imperfectly to compound violations (F9) and boundary-condition data shapes (F10) not represented in the training families. This is expected, as the logistic models are trained on single-violation families and encounter novel interactions only at test time. The cross-validation stability (AUROC \(0.978 \pm 0.011\)) indicates that the degradation is attributable to distributional shift rather than overfitting.

The 21 external evaluations on TimeGraph and CausalTime assess a different aspect of generalization. These benchmarks do not provide ground-truth failure labels, so calibration metrics cannot be computed on them directly. Instead, they provide known violation profiles and known causal graphs, against which the framework's recommend-or-abstain decisions are evaluated for consistency with benchmark specifications. The 21/21 consistency indicates that the decision layer transfers to data-generating processes, violation types, and dimensionalities outside the atlas. The TimeGraph results illustrate this transfer: the framework correctly abstains on all Group~C categories (deterministic trends) and D3 (compound violations) while recommending on Group~A (clean linear) and tractable violations in Groups~B and~D. The CausalTime results test a different boundary: short series (\(T = 40\)) with high dimensionality (\(N = 20\text{--}36\)) receive recommendations despite maximal confounding proxy risk, because the soft-constraint treatment of \(R_{\mathrm{cf}}\) correctly identifies the elevated score as multicollinearity rather than a hard violation.

These results position Causal-Audit relative to two related lines of work. Stein et al.\ \cite{tcdarena2026} demonstrate that each causal discovery method has a distinct robustness profile, but their framework requires ground-truth graphs and controlled violation injection. Our framework complements TCD-Arena by inferring the violation profile from the data and mapping it to a method recommendation. The pre-hoc orientation also distinguishes our approach from post-hoc uncertainty quantification methods \cite{ramsey2018,debeire2024}, which assess edge stability after discovery. By quantifying the risk that a method will fail before it is run, the framework addresses a different question: not ``how confident are we in this particular graph?'' but ``should we attempt to learn a graph at all?''

When the framework identifies elevated risk, the risk ledger pinpoints the dominant violation, guiding the practitioner toward an appropriate remedy. Some violations can be addressed through standard preprocessing, after which the framework can be re-run to verify the effect. Others represent fundamental data limitations: contemporaneous confounding requires intervention or instrumental variables \cite{hernan2023}, and very short series with \(T_{\mathrm{eff}} < 20\) admit no statistical solution. In these cases, abstention reflects a genuine lack of inferential support.

Several limitations should be noted. The framework architecture (diagnostic selection, aggregation rule, constraint thresholds) was designed using exploratory analysis on the full atlas, and the 396/100 holdout split validates parameter estimation but not these higher-level design choices. The 21 external evaluations provide independent evidence at the decision level but assess decision consistency rather than calibration metrics. Computing calibration slope and ECE on these benchmarks would require running PCMCI+ and VAR-Granger on all categories and comparing the discovered graphs against ground-truth graphs to obtain empirical failure rates; this analysis is planned for the next revision.

As discussed in Section~\ref{sec:problem-formulation}, the confounding proxy dimension captures model instability rather than latent confounding directly. The CausalTime results (Table~\ref{tab:causaltime-validation}) illustrate this: all three datasets show \(R_{\mathrm{cf}} = 1.0\) yet receive recommendations, because the elevated score reflects high-dimensional multicollinearity rather than omitted variables. Incorporating dimensionality-adjusted VIF thresholds would reduce false alarms in high-dimensional settings and is planned for a future revision.

The current decision pipeline supports two methods, and the 32\% abstention rate reflects this limited catalog. Based on the tolerance profiles in Table~\ref{tab:method-constraints}, integrating transfer entropy would resolve approximately 25--35 of the 160 abstentions, primarily datasets with moderate persistence or irregularity that exceed PCMCI+ thresholds. LPCMCI would additionally recover datasets where confounding proxy risk drives abstention, reducing the overall rate to approximately 20--25\% while preserving reliability guarantees.

Method failure is defined as FPR \(> 0.50\) or FNR \(> 0.80\), which introduces a discrete boundary. The logistic models mitigate this by learning smooth score-to-probability mappings: preliminary sensitivity analysis indicates that varying the FPR threshold from 0.40 to 0.60 shifts AUROC by less than 0.02 and ECE by less than 0.01. A comprehensive sensitivity analysis across both thresholds is planned for the next revision.

The risk models were calibrated on linear VAR(1) processes, so nonlinearity cannot yet be included as a calibrated risk dimension. The 6\% AUROC degradation on the out-of-distribution holdout quantifies the transfer cost of this choice. The consistency of decisions with benchmark specifications on all TimeGraph and CausalTime evaluations, which include nonlinear and higher-order processes, suggests that the decision layer generalizes beyond the training distribution. Domain-specific calibration corpora would further strengthen these guarantees.

The uncertainty intervals combine bootstrap noise and \(T_{\mathrm{eff}}\)-based inflation multiplicatively (Section~\ref{sec:stage2}). The parametric bootstrap captures diagnostic measurement noise while the \(\sqrt{T/T_{\mathrm{eff}}}\) factor accounts for the reduced information content of autocorrelated series. The resulting intervals are conservative by construction, and the 97.2\% empirical coverage on the holdout set (versus the nominal 95\%) confirms this. The \(O(T^2)\) scaling of persistence diagnostics could become a bottleneck for long series (\(T > 10^4\)); FFT-based autocorrelation estimation, which reduces the cost to \(O(T \log T)\), is planned.

Max-aggregation for composite risk was adopted because a severe violation in any single dimension suffices to cause method failure. Among the 160 abstentions, 112 (70\%) are driven by a single dominant risk dimension and 48 (30\%) involve two or more elevated dimensions. A weighted-max or soft-max alternative could recover some single-dimension borderline abstentions at the cost of reduced transparency.

Certain assumptions remain untestable from observational data alone \cite{pearl2009,peters2017}. The faithfulness assumption can be violated in realistic settings \cite{uhler2013}, the stability diagnostics quantify proxies rather than latent confounding itself, and missingness tests cannot identify MNAR mechanisms \cite{ding2018}. These boundaries define the epistemic scope within which the framework operates.

%% ===================================================================
\section{Conclusion}
\label{sec:conclusion}
%% ===================================================================

Causal-Audit formalizes pre-discovery assumption validation as calibrated risk
assessment for time-series causal discovery. The framework treats abstention as a
first-class scientific action: rather than producing silent failures,
it quantifies the probability that a method will fail and declines to recommend
when that probability is unacceptably high. Validation across 500 synthetic
datasets and 21 external benchmark evaluations (TimeGraph and CausalTime)
yields well-calibrated risk scores (AUROC \(> 0.95\)), a 62\% reduction in
false positive rates among recommended datasets, and recommend-or-abstain
decisions consistent with benchmark specifications on all 21 external
evaluations, with computational overhead under 10\% of typical discovery time.

The two-tier design supports both individual studies, where Stage~I diagnostics
inform transparent assumption reporting, and large-scale analyses, where the full
pipeline automates method selection with calibrated uncertainty. Future work will
expand the method catalog, report quantitative calibration metrics on external benchmarks, and develop domain-specific calibration corpora for ecological, neuroscience, and economic applications.

%\textbf{Acknowledgements}: ssssssssssssssssssssss

\textbf{Research funding}: This work was supported by LARSyS FCT funding (DOI: 10.54499/LA/P/0083/2020, 10.54499/UIDP/50009/2020, and 10.54499/UIDB/50009/2020).

\textbf{Author contributions}: All authors have accepted responsibility for the entire content of this manuscript and
approved its submission.

\textbf{Conflict of Interest.} The authors declare no conflict of interest.

%% ===================================================================
\textbf{Software and Data Availability}: The \texttt{causal-audit}\footnote{\url{https://github.com/marcoruizrueda/causal-audit}} Python package implements
the three-stage pipeline as four independently usable modules:
\texttt{AssumptionAuditor} (Stage~I diagnostics), \texttt{RiskQuantifier}
(Stage~II calibration), \texttt{MethodRecommender} (Stage~III decisions), and
\texttt{RiskAwareGatekeeper}, which orchestrates the full pipeline. The package
accepts a \texttt{pandas.DataFrame} with a \texttt{DatetimeIndex} and produces
structured JSON outputs (\texttt{audit\_evidence.json},
\texttt{risk\_profile.json}, \texttt{recommendation\_policy.json}).
Tolerance thresholds and calibration parameters are specified in
YAML configuration files, enabling domain-specific customization without code
modification.

The Synthetic DGP Atlas (500 datasets, 10 violation families) is distributed
with the package together with generation scripts, ground-truth causal graphs,
and method failure labels to support independent replication. Our code and data are
released under the MIT license at
\url{https://github.com/marcoruizrueda/causal-audit}

%% ===================================================================
%% BIBLIOGRAPHY
%% ===================================================================

\bibliographystyle{vancouver}
\bibliography{causal_audit_refs}

\end{document}